\documentclass[review]{elsarticle}

\usepackage[cm]{fullpage}
\usepackage{pdflscape} 
\usepackage{lmodern,amsmath,amssymb}
\usepackage{commath}
\usepackage{a4wide}
\usepackage{algorithm}
\usepackage{algpseudocode}
\usepackage{multirow}
\usepackage{graphicx}
\usepackage{multicol}
\usepackage{tabularx}
\usepackage{vwcol} 
\usepackage{color,soul}
\usepackage{subcaption}
\usepackage{url}
\usepackage{booktabs}
\usepackage{multirow}
\usepackage{mathtools}
\usepackage{placeins,cleveref}
\usepackage[export]{adjustbox}
\usepackage{placeins}
\usepackage{ragged2e}
\usepackage[toc,page]{appendix}
\usepackage{lipsum}
\usepackage{caption}
\usepackage{textcomp}
\usepackage{lscape} 
\newtheorem{theorem}{Theorem}[section]
\newtheorem{lemma}{Lemma}
\newtheorem{corollary}{Corollary}
\newtheorem{definition}{Definition}
\newtheorem{proof}{Proof}

\algnewcommand\algorithmicforeach{\textbf{for each}}
\algdef{S}[FOR]{ForEach}[1]{\algorithmicforeach\ #1\ \algorithmicdo}

\biboptions{square,sort}

\journal{Pattern  Recognition}

\begin{document}
	
	\begin{frontmatter}
		
		
		
		\title{\emph{IPD}:An Incremental Prototype based \emph{DBSCAN} for large-scale data with cluster representatives}
		
		
		\author{Jayasree Saha}
		\ead{jayasree.saha@iitkgp.ac.in}
		\author{Jayanta Mukherjee}
		\ead{jay@cse.iitkgp.ac.in}
		\address{Department of Computer Science and Engineering\\Indian Institue of Technology Kharagpur\\ West Bengal, India, 721302}
		
		\begin{abstract}
			\emph{DBSCAN}  is a fundamental density-based clustering technique that identifies any arbitrary shape of the clusters. However, it becomes infeasible while handling big data.  On the other hand, a centroid-based clustering algorithm, such as K-means,  for detecting patterns in a dataset since unprocessed data points can be labeled to their nearest centroid. However, it can not detect non-spherical clusters.  For a large data, it is not feasible to store and compute labels of every samples. These can be done as and when the information is required. The purpose can be accomplished when  clustering act as a tool to identify cluster representatives and query is served by assigning cluster labels of nearest representative.	In this paper, we propose an Incremental Prototype-based \emph{DBSCAN} (IPD) algorithm which is designed to identify arbitrary-shaped clusters for large-scale data. Additionally, it chooses a set of representatives for each cluster. 
		
		\end{abstract}
		
		
	\end{frontmatter}
	
	\section{Introduction}
 Clustering is the process of grouping similar objects into one cluster and dissimilar objects into separate clusters based on some similarity or dissimilarity functions. It is a well-known unsupervised tool used in several machine learning and data mining applications where ground truth is unavailable~\cite{ML_DM_PR09}.  However, the current era of Big Data has introduced new challenges to the existing machine learning and data mining approaches~\cite{big_data2020,ML_big2017}. Due to automatic capability of knowledge extraction, clustering becomes obvious choice for handling big data. \\
 Density based clustering is one of the most popular paradigm in the machine learning and data mining community. Ester et al. \cite{DBSCAN96} introduced Density-Based Spatial Clustering of Applications with Noise (\emph{DBSCAN}) which is a well-known density-based clustering algorithm. The idea is to group data in the high-density region of the feature space. 
 It requires two parameters: i) scanning radius $\epsilon$, and ii) a density threshold $MinPts$.
 It has the capability of recognizing clusters on complex manifolds, having arbitrary shapes. It is not limited to identifying only ``spherical" clusters as any centroid-based algorithm. However, the applicability of \emph{DBSCAN} on very large-scale datasets is limited due to its computational complexity   ~\cite{BLOCK-DBSCAN2021,fastdbscan2013}. Ester et al. claimed $\mathcal{O} (n log n)$ running time with a suitable index structure for data in d-dimensional Euclidean spaces, where $n$ is the number of objects. But, Gan and Tao \cite{DBSCANrevisited2015} recently proved that does not hold for $d > 3$, \emph{DBSCAN} requires at least ${\Omega}(n ^{\frac{4}{3}})$.   Algorithms with $\mathcal{O}(n)$ time complexity are still facing scalability issues to handle very
 big data where $n$ is counted in millions or billions. Hence, applying \emph{DBSCAN} on very big data is still a challenging task and needs a better solution.\\
 \noindent
 {\bf Contribution.} In this work, we focus on a prototype-based incremental clustering approach which {is completely built} upon a fraction of whole dataset. 
 We aim to identify cluster representatives instead of only providing partitions for the given dataset. The basic principle of the algorithm is based upon \emph{DBSCAN}.
 Our algorithm, called Incremental Prototype-based DBSCAN (\emph{IPD}) has following characteristics and benefits:\\
 \noindent
 1)  Data redundancy  is responsible for the volume of Big data~\cite{big_data_redundancy2012, big_data_redundancy2018}. Therefore, it is possible to summarize big data with  a fraction of the whole data. We have exploited this property in our clustering algorithm to make it suitable for big data. Usually, in the existing prototype-based approaches~\cite{l-dbscan2006, Rough-DBSCAN2009}, the prototype is first generated using leaders clustering algorithm~\cite{leader75} in linear time, followed by applying clustering algorithm on the prototypes. Since algorithms with $\mathcal{O}(n)$ time complexity are still facing scalability issues to handle very big data, sampling-based solutions are more attractive. 
 The existing techniques~\cite{AnyDBC2016, Sampling2019} requires $\mathcal{O}(n)$ to perform  each neighborhood query if no index structure is used. However, this complexity can be reduced  on the sample space since it reduces individual neighborhood query as well as total number of such queries. In this work, we focus on a sampling-based approach to build a prototype.  However, the suitability of the prototype is determined incrementally.\\
 \noindent
 \noindent
 2) Since data redundancy plays a vital role in the volume of big data, we restrict ourselves to process only a fraction of the dataset and identify the cluster structure.  This technique raises two questions: i) Which structural form is appropriate for representing a cluster structure?, and ii) How can we validate that the cluster structure obtained from the sample space is a good fit for the whole dataset?\\ 
 \noindent
 i) In partition-based methods (e.g, \emph{K-means} ~\cite{kmeans_Lloyd57}), a cluster is represented by its centroid. \emph{K-means} is mostly suitable for convex clusters. However, a single member cluster representation scheme is not suitable for labeling data points in the case of closely spaced arbitrarily shaped clusters.
 According to Tong et al.~\cite{boundary2017} boundary points can be potential candidates for representing a cluster in a prototype-based clustering algorithm. Hence, we have exploited the benefits of having boundary points as  representatives beside the centroids.\\ 
 \noindent	
 ii) If the clustering structure does not change while perturbing the corresponding sampled dataset, we may consider that the clustering structure is stable. In the current context, we introduce perturbation by adding new data points into the sample iteratively. We examine whether the clustering structure changes in every iteration. The clustering structure of the sampled dataset resembles the cluster structure of the whole dataset when it accomplishes the stability. \\

 \noindent
 {\bf Summarization.} Our major contributions are as follows:
 
 \begin{itemize}
 	\item We introduce a density based clustering method which is based on the basic principle of \emph{DBSCAN}. It  uses a fraction of the whole dataset to produce a suitable cluster structure.  Hence, it reduces each query processing time and the total  number of queries. The main benefits of our algorithm is that it identifies cluster structure in terms of cluster representatives. In real time, one may not be interested in knowing the labels of all data points  but a few for large scale data. Our strategy makes the algorithm more robust and efficient for handling large scale data in real time scenario.
 	
 	\item We introduce an incremental scheme to identify cluster structures by employing a stability criterion.  Our technique relies on the sampled items on each iteration  for querying. Hence, it reduces the number of queries. 
 	
 	\item Our method introduces a strategy for having multiple representatives  for each cluster. It facilitates a new instance to be classified  to a cluster having any arbitrary shape using the nearest neighbor rule. This labeling scheme is more robust and accurate compared to the single representative scheme as in partition based methods. 
 \end{itemize}

 To the best of our knowledge, our work is the first work which incorporates the idea of having representatives in density based clustering algorithm. To demonstrate the scalability and efficiency of our algorithm, we conduct experiments on synthetic and real-world datasets where size varies from $10^3$ to $10^6$. \\
 The rest of the paper is organized as follows: we introduce related works in \Cref{sec:background} , and then detail the 
 \emph{DBSCAN} in \Cref{sec:dbscan}. \Cref{sec:methodology} presents our proposed algorithm in details. Evaluation results are given in
 \Cref{sec:results}. 
 \begin{table*}[htb!]
 	\centering
 	\caption{{Notations used in the proposed work}}
 	\begin{tabular}{clcl}
 		\hline
 		\textbf{Notation} & \multicolumn{1}{l}{\textbf{Definition}}                                                                                               
 		\\ \hline
 		$S_{prototype}$   & \begin{tabular}[c]{@{}l@{}}A sample chosen for prototype creation s.t. $|S_{prototype}|=\gamma$\end{tabular}                              \\ $S_{test}$                            & \begin{tabular}[c]{@{}l@{}}A sample chosen for definition the notion of stability s.t. $|S_{test}|=\alpha$\end{tabular}                     \\ 
 		$S_{inc}$         & \begin{tabular}[c]{@{}l@{}}A sample chosen during incremental  processing s.t. $|S_{inc}|=\beta$\end{tabular} \\                             $C$                                   & A cluster                                                                                                                               \\ 
 		$\mathbb{C}$       & A set of clusters   \\                                                                                                                   $\epsilon$                     &Scanning radius (\emph{DBSCAN} parameter) \\
 		$ MinPts$                     & Density threshold to become a core point (\emph{DBSCAN} parameter)                                                                                                                           \\ 
 		$K$               & Total number of clusters    \\                                                                                                           $G_{oracle}$                          & \begin{tabular}[c]{@{}l@{}}A graph obtained when DBSCAN is applied on $X$\end{tabular}                                                \\ 
 		$G_{prototype}$   & \begin{tabular}[c]{@{}l@{}}A graph obtained when DBSCAN is applied on $S_{prototype}$\end{tabular} \\                                  $G^{\prime}_{prototype}$               & \begin{tabular}[c]{@{}l@{}}A graph obtained when items in $S_{inc}$ is  included in $G_{prototype}$\end{tabular}                      \\ 
 		 $\gamma$ & \begin{tabular}[c]{@{}l@{}}Size of initial prototype  in IPD.\end{tabular}  \\
 		$\beta$ & \begin{tabular}[c]{@{}l@{}}Size of incremental sample.\end{tabular}  \\
 		$\alpha$ & \begin{tabular}[c]{@{}l@{}}Size of test sample.\end{tabular}  \\
 		$N_{\epsilon}(p)$     & \begin{tabular}[c]{@{}l@{}}A set of points in $\epsilon$-neighborhood of  point $p$.\end{tabular}                        \\ $\eta$                                 & \begin{tabular}[c]{@{}l@{}}It controls the value of $MinPts$ in sample space in \emph{IPD}\end{tabular}                                                        \\ 
 		$\mathcal{R}$                          & \begin{tabular}[c]{@{}l@{}}A set of representatives of a cluster partition\end{tabular}                                               \\ 
 		$\tau$ & Threshold for selecting representatives.\\
 		$\Omega_{test}$    & \begin{tabular}[c]{@{}l@{}}Cluster labels for $S_{test}$ at timestamp $t$\end{tabular}        \\                                      $\Omega^{\prime}_{test}$                & \begin{tabular}[c]{@{}l@{}}Cluster labels for $S_{test}$  at timestamp $t+1$\end{tabular}                                             \\
 		$\varDelta$       & Stability metric                                                                                                                                                                                                                                                                                                    \\ \hline
 	\end{tabular}
 	\label{tab:notations}
 \end{table*}
\section{Related work}\label{sec:background}

\subsection{Variants of \emph{DBSCAN} for big data }
\emph{DBSCAN} {can detect arbitrary shaped clusters depending on chosen $\epsilon$ and $MinPts$.} But, it runs with quadratic time complexity. Therefore, the time requirement of this technique becomes intolerable in the era of big data. Several variants of \emph{DBSCAN} evolved in decades to make it usable for large-scale and high-dimensional data. 
Tong et al.~\cite{boundary2017} incorporate  Scalable Clustering Using Boundary Information (\emph{SCUBI}) into \emph{DBSCAN}. The idea of \emph{SCUBI} is to identify the boundary points of the original dataset followed by grouping them into suitable clusters using any clustering techniques. Finally, the
remaining points receive  cluster labels from those of their nearest boundary points.

The basic approach of a hybrid clustering technique~\cite{l-dbscan2006, Rough-DBSCAN2009} is to find suitable prototypes from the large dataset and apply the clustering algorithm on the prototypes. In literature, the leaders clustering algorithm~\cite{leader75} derives prototypes in linear time.  According to Viswanath et al.~\cite{Rough-DBSCAN2009}  leaders alone cannot be used to obtain the density information. {Therefore, followers of the leaders is also considered for the prototype.}
Such schemes provide an approximate solution that may deviate from the methods which consume the entire dataset. However, such deviation depends upon the quality of the prototypes.  {Additionally, run time complexity of such algorithm is $\mathcal{O}(n)$ which could be a bottleneck for handling big data.}

The principle of Grid-based DBSCAN~\cite{grid2004, grid2013} is to divide the original dataset into equal-sized square-shaped grids where any two objects in the same grid belong to the  $\epsilon$-neighborhood of each other. The Grid-based DBSCAN uses neighbor grid queries instead of $\epsilon$-range queries and merges grids. 
However, grid techniques are usually limited to high dimensional data space due to two factors: 1) neighbor explosion, and 2) a substantial amount of redundant distance computations during merging. 
Several improved algorithms in the literature try to minimize these factors.
For example, Bonchoo et al.\cite{grid2019} employ bitmap indexing to provide efficient neighbor grid queries. They incorporate a forest-like structure to alleviate the redundancies in the merging. 
Chen et al.~\cite{BLOCK-DBSCAN2021} introduced  \emph{BLOCK-DBSCAN} where $\frac{\epsilon}{2}$-norm ball  finds several core points at one time within  ``Inner Core Blocks." Then, two such blocks are merged if they are density reachable. 

Besides, there are  other \emph{DBSCAN} variants for handling big data.  AnyDBC~\cite{AnyDBC2016} is an anytime DBSCAN which
employs an active learning scheme to a subset of data points for refining cluster structure iteratively.  Mai et al. \cite{IncAnyDBC2020} introduces {\emph IncAnyDBC}, which processes neighborhood queries in a block in every iteration to build clusters. They utilized this scheme to parallelize the algorithm in shared memory structures such as multicore CPUs. 

A few  methods~\cite{cudaSCAN2015,CUDADClust2009} utilize Graphics Processing Units (GPUs) to improve the processing speed of the conventional algorithm. The major challenge in GPU is due to the repeated computation of distances from all objects. However,  storing the data points and intermediate clustering results in the off-chip memory of the GPU is very costly. Some methods~\cite{MR-DBSCAN2011, NG-DBSCAN2016} use map-reduce algorithms to exploit distributed and parallel architecture. 

\subsection{Notion of stability in Clustering}
Cluster analysis is governed  by two factors: 1) whether a data is clusterable~\cite{ADOLFSSON2019_clusterability}, and 2) whether the clustering results are stable~\cite{VOLKOVICH2008_cluster_stability, Alessandro2012_density_stability}. 
Clusterability aims to quantify the degree of cluster structure, and the checking should happen before applying any clustering algorithms.  With proper tuning of parameters, \emph{DBSCAN} may detect unimodality, but there is no statistical test embedded into the algorithm to justify its correctness. On the other hand, stability is measured only after getting a partition. In the current context, we focus on the second factor. We aim to provide a clustering algorithm that can provide a stable partition. Therefore, we discuss the current trends in clustering stability.  \\
\noindent
The notion of stability usually assesses the variability of clustering under small perturbation in the data~\cite{K_stbility_ben02}. In general, stability is measured by computing the distribution of pairwise similarities or dissimilarities between clusterings obtained from subsamples of the data. The concept of stability is explored for checking cluster validity~\cite{roth_stability} as well as determining the true cluster number~\cite{instability2012, K_stability_clest} associated with the dataset. Wang \cite{instability_Wang10}  shows a way to determine a valid number of clusters applicable for the dataset via cross-validation. The author first divides the whole dataset into three sets randomly. Two of them are used for training two clustering models and another set is used for validation. The validation set is labeled by two models for a given cluster number $K$. Then, it counts the number of pairs of elements in the validation dataset, which are labeled differently by two models. The process is repeated several times and it computes the average of that count. The average acts as the instability for the value $K$ on the dataset. They repeat the process for the series of cluster numbers. The valid cluster number is the one which is having minimum instability. In this paper, we utilize the concept of instability to assess the validity of cluster structure in the incremental approach. However, designing cluster validation criteria in a sample-based framework~\cite{NIPS2007_Stability_Finite_Samples} is a challenging task. The main challenge is to identify meaningful clustering and assure that the clustering does not correspond to any artifacts of the sampling process. 
In particular, clustering solutions become more stable as the sample size increases.
Shamir et al.~\cite{NIPS2008_Reliability_Clustering_Stability} studied a consistency condition, a central limit condition, and
regularity conditions as the general sufficient condition to ensure the reliability of clustering stability estimators in the substantial sample regime.

\section{Preliminaries}~\label{sec:dbscan}
\emph{DBSCAN}~\cite{DBSCAN96} is one of the popular density based clustering algorithms which is capable of capturing clusters with arbitrary shape and size. It requires two parameters $\epsilon$ and $MinPts$. $\epsilon$ is the radius of the neighborhood of a point. $\text{MinPts}$ set the threshold for a point to become a core point. To obtain a cluster, \emph{DBSCAN} finds all points density-reachable from any point $p$ wrt. $\epsilon$ and $\text{MinPts}$. We further discuss some important concepts and terms used in \emph{DBSCAN}.

\subsection{Basic concepts}~\label{sec:basic}
Let $X = \{x_1,\cdots x_n\}$ be the original dataset, where  $x_p\in \mathcal{R}^d$ represents the $p^{\text{th}}$ object.
$dist(x_i, x_j)$ denotes the distance between $x_i$ and $x_j$.\\
\noindent
\begin{definition}
	The {\bf $\epsilon$-neighborhood } of a point $p \in X$ , denoted by $N_{\epsilon}(p)$, is defined as, $N_{\epsilon}(p)=\{q|q \in X$ and $dist(p,q)\leq \epsilon\}$ 
\end{definition}
\begin{definition}
	A point $p$ is a {\bf core point} if $|N_{\epsilon}(p)| \geq \text{MinPts}$	
\end{definition}

\begin{definition}
	A point $p$ is {\bf directly density-reachable} from $q$ w.r.t. $\epsilon$ and  $\text{MinPts}$  if $p \in N_{\epsilon}(q)$ and $q$ is a core point. We use the notation $q \leftrightarrow p$ if $p$ is directly density-reachable from $q$. It also implies that $q$ is directly-reachable from $p$.
\end{definition}

\begin{definition}
	A point $p$ is a {\bf border point} when $|N_{\epsilon}(p)| < \text{MinPts}$ and  $\exists q \in N_{\epsilon}(p)$ such that  $q$ is a core point.
\end{definition}

\begin{definition}
	A point $p$ is a {\bf noise point} when $|N_{\epsilon}(p)| < \text{MinPts}$ and  $\forall q \in N_{\epsilon}(p)$,  $q$ is a not a core point.
\end{definition}

\begin{definition}
	A point $p$ is {\bf density-reachable} from $q$ if there is a series of points $p_1, p_2,\cdots, p_m$ such that $p_1=p$,  $p_m=q$ and $p_{i+1}$ is directly density reachable from $p_{i}$ where $1 \leq  i \leq m - 1$
\end{definition}

\begin{definition}
	A {\bf cluster} $C$ with respect to $\epsilon$ and $MinPts$ is the maximal set of density connected points. Noise points do not belong to any cluster.
\end{definition}

\begin{definition}
	A {\bf Graph}  $ \boldmath  G_\text{\bf oracle} (\mathcal{V}, \mathcal{E})$ with respect to $\epsilon$ and $MinPts$ represents the cluster structure $\mathbb{C}=\{C_1,\cdots,C_K\}$ produced by \emph{DBSCAN} where $|\mathcal{V}|=n$, and for every $e(u, v) \in \mathcal{E}$, $u \in N_\epsilon(v)$. Each maximal set of density connected vertices form a cluster $C_k$.
\end{definition}

\subsection{{DBSCAN} algorithm}
{\emph DBSCAN} selects randomly a point $p \in X$. It checks neighborhood of $p$ with respect to $\epsilon$. If cardinality of $N_{\epsilon}(p)$ is {larger than the threshold} $\text{MinPts}$, $p$ is a ``core" point and becomes a member of the cluster $C$.  Then, cluster $C$ is expanded by all the points which are density reachable from $p$. However, if $p$ is  not a ``core" point,  it is denoted as ``noise" temporarily. Then, {\emph DBSCAN} selects randomly another point $q$ which is not processed in the previous step and assigns  cluster $C+1$, if it is a core point. Subsequently, the cluster $C+1$ is expanded. If any point which is density reachable from $q$ but categorized as a noise in the previous step, it is marked ``border" point.  The algorithm stops when all the points in the database are processed. 
Algorithm~\ref{algo:dbscan} and Algorithm~\ref{algo:expand} together describe the \emph{DBSCAN} algorithm.

\begin{algorithm}[htb!] 
	\caption{ DBSCAN ($X$, $\epsilon$, $\text{MinPts}$)} 
	\label{algo:dbscan}
	
	
	\begin{algorithmic}[1]
		\State Initialize Cluster id $C=0$
		\ForEach{$p \in X$}
		\If{\emph{ExpandCluster($p, \epsilon, \text{MinPts}, C$)}} 
		\State $C=C+1$
		\EndIf
		\EndFor	
	\end{algorithmic}
\end{algorithm}

\begin{algorithm}[htb!] 
	\caption{ExpandCluster ($p$, $\epsilon$, $\text{MinPts}$, $C$)} 
	\label{algo:expand}
	
	
	\begin{algorithmic}[1]
		\State $N_{\epsilon}(p) \leftarrow$ \emph{RangeQuery}($p, \epsilon$)
		\If{$|N_{\epsilon}(p)|< \text{MinPts}$}
		\State Mark $p$ as {\emph ``noise"} 
		\State return {\bf False}
		\Else
		\State label $t \in N_{\epsilon}(p)$ with $C$
		\State $seeds \leftarrow N_{\epsilon}(p)$
		\EndIf
		
		\ForEach{$q \in {seeds}$}

		\State $N_{\epsilon}(q) \leftarrow$ \emph{RangeQuery}($q, \epsilon$)
		\If{$|N_{\epsilon}(q)| \geq$ MinPts } 
		\State Mark $q$ as {\emph ``core"} 
		
		\ForEach{$r \in N_{\epsilon}(q)$}
		\State label $r$ with $C$
		\If{($r$ is not processed)  or ( $r$ is a {\emph ``noise"})}
		\If{$r$ is not processed}
		\State $seeds= seeds \cup r$
		\Else
		\State Mark $r$ as {\emph ``border"} 
		\EndIf
		
		\EndIf
		\EndFor	
		
		\EndIf
		
		\EndFor	
		\State return {\bf True}
		
	\end{algorithmic}
\end{algorithm}

\noindent
The function \emph{RangeQuery}($p$, $\epsilon$) returns all neighbors within the $\epsilon$-neighborhood of $p$. 
The time complexity of {\emph DBSCAN} depends upon the running time of the  \emph{RangeQuery}($p, \epsilon$) which must be performed for each point.  Intuitively, it requires $\mathcal{O}( n^2 )$ unless any indexing scheme is used.

\section{IPD}\label{sec:methodology}
The general ideas of \emph{IPD} includes:
{creation of a prototype for the cluster structure present in $X$}, incremental processing,  {ensuring stability for a set of clusters}, and  generation of cluster representatives. The pseudo-code for \emph{IPD} is summarized in \Cref{algo:IPD}. \Cref{tab:notations} shows notations used in the paper.

\subsection{Prototype creation}
We first choose a sample  $(S_{prototype})$ from the list of unprocessed points in the original dataset as described in \Cref{def:proto}. 

\begin{definition}\label{def:proto}
	We define a sample $S_{\text{prototype}}$ 
	such that $S_{\text{prototype}} \subset X$ and each object in $S_{\text{prototype}}$ is sampled randomly with i.i.d and without repetition, such that $|S_{\text{prototype}}|=\gamma$. The remaining points are $X_{\text{rem}} =X-S_{\text{prototype}}$.  $S_{\text{prototype}}$ acts as initial prototype for the clustering
\end{definition}
\noindent
Then, we apply  {\emph DBSCAN}~\cite{DBSCAN96}  on this sample. i.e., the query of a point only searches $\epsilon$-neighborhood in the sample space but not in the original data space. 
\begin{figure}[htb!]	
	\centering
	\begin{multicols}{2}
		\subcaptionbox{DBSCAN\label{fig:dbscan}}{\includegraphics[width=0.35\textwidth]{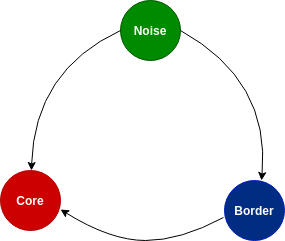}}

		\subcaptionbox{IPD\label{fig:riscan}}{\includegraphics[width=0.35\textwidth]{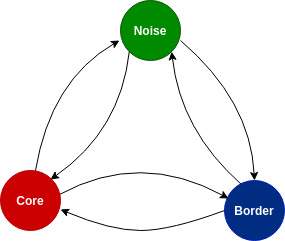}}

	\end{multicols}
	\caption{The state transition diagram of a point}
	\label{fig:transition}
\end{figure}
\noindent
Each point undergoes state transition during the execution of \emph{DBSCAN} as shown in \Cref{fig:dbscan}. Intuitively, the size of  $\epsilon$-neighborhood of a point in the sample space is small compared to the original data space. Therefore, the value of the parameter $MinPts$ in \emph{DBSCAN}  may not be appropriate to identify the ``core" point when \emph{DBSCAN} applied in the sample space. To resolve the issue, we introduce parameter $\eta$ to control the value of $MinPts$ in the sample space. This scheme reduces the computation time for neighborhood queries and builds the structure of a tentative cluster on sample space.
In this step, we create a graph $G_{p}(V, E)$ as defined in \Cref{def:proto_graph}, based on the outcome of \emph{DBSCAN}. 

\begin{definition}\label{def:proto_graph}
	We define a graph $G_\text{prototype}=(V, E)$ such that $v \in V$ 		represents a point in $S_{\text{prototype}}$.  We have tagged each vertices with any of three states: ``core", ``border", and ``noise". The definitions of these states have the same implications as discussed in \Cref{sec:basic}. These states are decided when \emph{DBSCAN} is applied on $S_{\text{prototype}}$. For every $e(u,v) \in E$, $u$ is in $\epsilon$-neighborhood of $v$. We assign ``volatile-yes" (v-yes), ``volatile-weak"(v-weak), and ``volatile-no"(v-no) state to an edge $e(u,v)$ if $u$ and $v$ are both core, if only one of them is core, and  both are not core, respectively. Each maximal set of ``volatile-yes" and ``volatile-weak" connected vertices forms a  cluster.
\end{definition}
\noindent
We  store the neighbors for each point $m$ (a vertex in $G_\text{prototype}$) , denoted as $N_{\epsilon}(m)$ , for determining the core property of $m$. This data structure is extremely useful in the incremental step to avoid unnecessary queries of processed points.  Since the state of each vertex $v$ in $G_\text{prototype}$ may change in the subsequent steps, a few lists throughout the algorithm are maintained. These lists are called list of \emph{state} ({LOS}), list of \emph{core} ({LOC}) , list of \emph{border} ({LOB}), and list of \emph{noise} ({LON}). LOS denotes the current state of a point. The length of the LOS is $n$.
This list maintains the current state of a point. They are labeled to ``unknown" state at the beginning of the algorithm. \emph{DBSCAN}  changes the state of a few points ( points present in $S_{prototype}$) which is reflected in the \emph{State} list. {\em LOC, LOB}, and {\em LON} store points which are currently core, border and noise respectively. These are temporary lists whose sizes change several times in the life span of \emph{IPD}.

\subsection{Incremental processing}
In this step, we randomly sample $\beta$ points from the remaining dataset. We update the existing cluster structure while processing them. 

\begin{definition}\label{def:sinc}
	We define sample ${S_\text{inc}}$ where each object is sampled randomly with i.i.d and without repetition from $X_{\text{rem}}$ such that $|{S_\text{inc}}|=\beta$. We modify $G_\text{prototype}=(V, E)$ to $G^{\prime}_\text{prototype}(V^{\prime}, E^{\prime})$ such that  $V^{\prime}=V\cup V_1$, where $v \in V_1$ represents a point in $S_{\text{inc}}$ and $E^{\prime}=E\cup E_1$, where $e(u, v) \in E_1$ such that $u\in{V^{\prime}}$, $v \in{V_{1}}$ and dist$(u, v)<=\epsilon$.  Also, we allocate state to edges as described in Definition 10. 	
\end{definition}
%
%
%
%
%
%
%
\noindent
\begin{figure*}[htb!]
	\centering
	\centering
	\begin{multicols}{3}
		
		\subcaptionbox{ \emph{Core}$\rightarrow$\emph{Core}\label{fig:core2core}}{\includegraphics[width=0.32\textwidth]{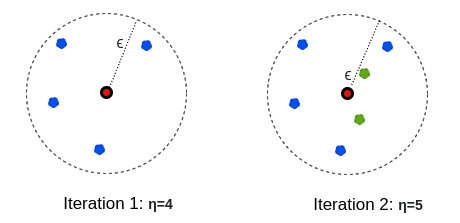}}

		\subcaptionbox{\emph{Core}$\rightarrow$\emph{Border}\label{fig:core2border}}{\includegraphics[width=0.33\textwidth]{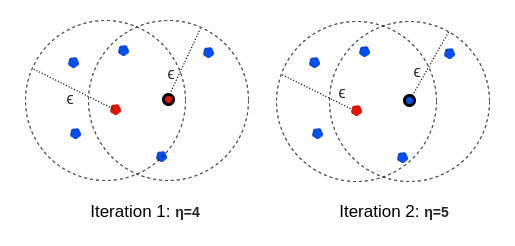}}
		
		\subcaptionbox{\emph{Core}$\rightarrow$\emph{Noise}\label{fig:core2noise}}{\includegraphics[width=0.3\textwidth]{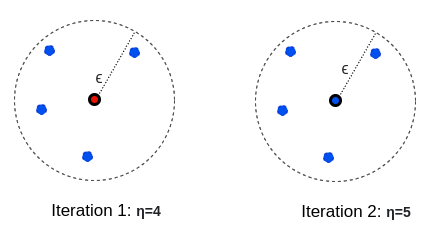}}

	\end{multicols}
	\caption{Transition of the core property of a point outlined with black color. circle represents $\epsilon$-neighborhood area, blue and green points are non core points, and red is the core point. Green points are added in the second iteration of incremental processing.  (a)  depicts the scenario when new points are added in the neighborhood of a core point in the next iteration. (b) and (c)  depict the situation when there is no increment in the neighborhood of a core point in the next iteration.}
	\label{fig:core_state_transition}
\end{figure*}

\noindent
Since we add more points into the prototype, the size of the $\epsilon$-neighborhood of a point in the sample space may increase. Hence, we increase the value of $\eta$ successively until $\eta=MinPts$. With this scheme, the core property of a point incurs several transitions as depicted in \Cref{fig:core_state_transition}. \\
When $MinPts$ is large, the number of iterations in incremental processing increases. To reduce such complexity, we sampled a fraction of the prototype. We observe the number of neighbors within $\epsilon$ of the sampled data points and take their mean.  We use this number to increment of $\eta$ till $\eta<MinPts$. Otherwise, we increase $\eta$ by $1$. \Cref{algo:EIM} describes the process.\\
\begin{algorithm}[htb!] 
	\caption{\emph{\small
			Estimation of  Incremental  MinPts ({EIM})  }}
	\label{algo:EIM}
	\small
	{\bfseries Input : }{$X, \epsilon, Minpts, \eta$}\\
	{\bfseries Output : }{$\eta$}
	\begin{algorithmic}[1]
		\State $\mathcal{S}_{i} \leftarrow$ sample a $x\%$ of  prototype randomly without repetition fron $X$.
		\State $\mathcal{J}=[]$
		\ForEach{$p \in \mathcal{S}_i$}
		\State $\mathcal{J}=\mathcal{J} \cup$ the number of $\epsilon$-neighbors of $p$
		\EndFor
		\State $\eta_t\leftarrow mean(\mathcal{J})$
		\If{$\eta_t>Minpts$}
		\State	$\eta \leftarrow \eta +1$
		\Else
		\State	$\eta \leftarrow \eta_t$
		\EndIf
	\end{algorithmic}
\end{algorithm}	
%
%
%
%
%
%
\noindent
In the incremental step merging of clusters can occur along with the addition of new cluster.  If the new point is a core point and its neighborhood contains points that belong to different clusters, then algorithm triggers merging of those clusters. \\
\noindent
\noindent

\begin{algorithm}[htb!] 
	\caption{\emph{\small Re-evaluation of ``Core" ({RC})}}
	\label{algo:updateCore}
	\small
	{\bfseries Input : }{List: $core$, $state$; Parameter: $MinPts$;
		Neighborhood map: $N_{\epsilon}$}\\
	{\bfseries Output : }{List: $core$,  $state$}
	\begin{algorithmic}[1]
		\ForEach{$x \in core$}
		\If{$|N_{\epsilon}(x)|<MinPts$}
		\State Remove $x$ from $core$
		\If {\emph{isNoise}(x)}
		\State $label(x) \leftarrow -1$
		\State $state(x) \leftarrow $``noise"\Comment{``core" $\rightarrow$ ``noise"  }
		\Else
		\State $state(x) \leftarrow $``border" \Comment{``core" $\rightarrow$ ``border" }
		\State $label(x) \leftarrow $ label of any core in its $\epsilon-$neighborhood.
		\EndIf 
		\State Update $state(y)$ and $label(y)$ for  any further state changes as in \Cref{lemma:cascadingEffect}. \Comment{$y \in N_{\epsilon}(x)$}
		\EndIf
		\EndFor	
	\end{algorithmic}
\end{algorithm}	

{\bf Re-evaluation of ``core" property.} 
At every iteration, we increment the value of $\eta$ that judges the ``core" property. Therefore, core property needs to be re-evaluated before processing $S_\text{inc}$. The \Cref{algo:updateCore} describes the process. 
\begin{algorithm}[htb!] 
	\caption{\emph{\small Re-evaluation of ``Noise" and ``Border" ({RNB})}}
	\label{algo:updateNB}
	\small
	{\bfseries Input : }{List: $noise$, $state$; Parameter: $MinPts$;\\
		Neighborhood map: $N_{\epsilon}$, Point: $x$}\\
	{\bfseries Output : }{List: $core$,  $state$}
	\begin{algorithmic}[1]
		\If{$state(x)=$``noise"}
		\State $label(p) \leftarrow \hat{k}$
		\If{$N_{\epsilon}(p)>=MinPts$}  \Comment{(noise $\rightarrow$ core) }
		\State $state(x)=$``core" 
		\State $core = core \cup x$
		\Else \Comment{(noise $\rightarrow$ border )}
		\State $state(x)=$``border" 	
		\EndIf
		\ElsIf{$state(x)=$``border"} \Comment{border$\rightarrow$ ``core"}
		\If{$N_{\epsilon}(p)>=MinPts$}
		\State $state(x)=$``core" 
		\State $core = core \cup x$
		\EndIf	
		\EndIf	
	\end{algorithmic}
\end{algorithm}	

\begin{lemma}\label{lemma:cascadingEffect}
	With the increment of $\eta$, if the state of a vertex $u$ looses  ``core" property, it may induce many state changes of points in its $\epsilon$-neighborhood. 
\end{lemma}

\begin{proof} 
	Let $u$ be chosen as a ``core" point when $\eta=t$ and $|N_{\epsilon}(u)|=t$. In the next iteration $\eta$ becomes $t+r$ ($r\geq1$) and $u$ becomes either ``border" or ``noise". $\exists v \in N_{\epsilon}(u)$ whose state is ``border" and $u$ is the only ``core" point in its $\epsilon$-neighborhood. i.e, $|N_{\epsilon}(v)|<t$ and there is no ``core" point in the $\epsilon$-neighborhood of $v$. Therefore, $v$ becomes  a ``noise". {Each core $\rightarrow$ border transition may induce border $\rightarrow$ noise transition.} ~~~~~~~~$\square$
\end{proof} 
\begin{lemma}\label{lemma:nocascadingEffect}
	There is no {further state changes} for border and noise points with the increment of $\eta$ unlike core points.
\end{lemma}
\begin{proof}
	Let $u$ be chosen as a ``border" point when $\eta = t$ and $|N_{\epsilon}(u)| = t-m$ where $1 \leq m < t$. When $\eta$ becomes $t+1$ in the next iteration, the size of $|N_{\epsilon}(u)|$ remains the same and the state is also not changed. Therefore, $u$ can not affect others in its $\epsilon$-neighborhood. The same situation holds for a ``noise" point. Therefore, there is no further state changes only with the increment $\eta$.~~~~~~~~$\square$
\end{proof}
\FloatBarrier
\begin{algorithm}[htb!] 
	\caption{\emph{\small incDBSCAN}}
	\label{algo:incDBSCAN}
	\small
	{\bfseries Input : }{Unprocessed Sample: $S_\text{inc}$\\
		Parameters for incremental DBSCAN: $\epsilon, MinPts$\\ New Cluster Id: $\hat{k}$}\\
	Lists: $state$ , $core$, $border$, $noise$ \\
	{\bfseries Output : }{Cluster Labels : $\mathbb{C}$, List : $\emph{State}$, Next Cluster Id: $\hat{k}$}
	\begin{algorithmic}[1]
		\For{$p \in S_\text{inc}$}
		\If{$p$ is not processed}
		\State	$Found, k \leftarrow$ \emph{incExpandCluster}($p,\epsilon, MinPts, \hat{k}$)
		\EndIf
		\If{ ($Found ==True$) and ($ k==\hat{k}$)}
		\State $\hat{k} \leftarrow \hat{k}+1$
		\EndIf
		
		\EndFor
		
		\State \Return $\hat{k}$
	\end{algorithmic}
\end{algorithm}	
\noindent
{Since the ``border" or ``noise" point does not induce further state changes, new points can be queried without changing their actual state when $\eta$ is incremented.} However, the size of $\epsilon$-neighborhood of $v \in S_\text{prototype}$ may increase while we run queries for every point in $S_{inc}$. Hence, we can re-verify ``border" or ``noise" property of $v$ while processing $S_\text{inc}$.\\ 
{\bf {Update the prototype graph.}}
During this step, the graph $G_\text{prototype}$ is modified so that it includes all merging as a result of new queries prompted by new samples $S_{\text{inc}}$ and new clusters within the new samples. 
At each iteration, \emph{IPD} randomly chooses a set ($S_\text{inc}$) of $\beta$
points from the remaining unprocessed dataset and run queries among their neighbors in $S_\text{prototype}$ and $S_\text{inc}$.  With state changes, $G^{\prime}_\text{prototype}$ may produce variation in cluster structure in $G_\text{prototype}$. 
\noindent
The cluster structure  $\mathbb{C}$ of $G_\text{prototype} (V, E)$ is updated to $\mathbb{C}^{\prime}$ of $G^{\prime}_\text{prototype}(V\cup V_1 ,E^{\prime})$ whose formation leads to the following conditions:
\begin{enumerate}
	\item If $\exists v \in V_1$ such that dist$(u, v)<=\epsilon$ where $u\in V$ and $u$ has a ``core" state and belong to $C_m$ cluster, then $v$ also belongs to  $C_m$ cluster. 

	\item If $\exists v \in V_1$ and $v$ has a ``core" state and $\exists u\in N_\epsilon(v)$  
	such that $u\in V$ and has ``noise" state. Then, maximal set of ``v-yes" and ``v-weak" connected vertices of $v$ is assigned to a new cluster $C_p \notin \mathbb{C}$ and  $v$'s state is changed to ``border" state. 

	\item If $\exists v \in V_1$ and $\epsilon$-neighborhood of $v$ contains core vertices $u$ and  $w$ such that $u\in C_m$ and $w\in C_n$ where $C_m\neq C_n$, then $C_m$ and $C_n$ are merged to $C_m$ $(m<n)$ and $v$ with its maximal set of ``v-yes" and ``v-weak" connected vertices (belong to $S_\text{inc}$) have been assigned to $C_m$.

\end{enumerate}


\noindent
\begin{lemma}\label{lemma:density_reachability}
	If two core points $u$ and $v$ are directly density connected
	such that $u \in S_\text{prototype}$, $u\in c_1$,  and $v \in S_\text{inc}$, there exists a path of vertices in $G^{\prime}_\text{prototype}$ that
	connects $\forall{w} \in c_1$ and $v$. Similarly, $\forall{z} \in S_{inc}$ if $z$ is density reachable to $v$, then $z$ is density connected to $u$ and $z$ belongs to the cluster $c_1$.
\end{lemma}

\begin{proof}
	Let $u \leftrightarrow x_1 \leftrightarrow  x_2 \cdots \leftrightarrow x_m \leftrightarrow w$ be a chain of core points connecting $u$ and $w$ (Definition 6, 10). After performing {query} on $v$, if $u \in N_{\epsilon}(v)$, then $dist(u,v)\leq \epsilon$. Therefore, $v \leftrightarrow u \leftrightarrow x_1 \leftrightarrow  x_2 \cdots \leftrightarrow x_n \leftrightarrow w$ i.e, $v$ is density reachable from $w$. Similarly, $\forall{z} \in S_{inc}$, if $z \leftrightarrow y_1 \leftrightarrow  y_2 \cdots \leftrightarrow y_p \leftrightarrow v$, then after querying on $v$, $z$ becomes density reachable to $u$.  $z\cdots \leftrightarrow v  \leftrightarrow u \cdots \leftrightarrow w$ forms a subset of the maximal set of density connected points. Hence, $z\in c_1$ according to Definition 7. ~~~~~~~~$\square$
\end{proof}
\noindent
Similar to \Cref{lemma:density_reachability}, if  $v \in S_\text{inc}$ and $u, w \in S_\text{prototype}$ but $u\in c_m$ and $w\in c_n$ such that $c_m \neq c_n$, then two clusters merge to a single cluster since $u \leftrightarrow v \leftrightarrow w$.
\Cref{algo:incDBSCAN} and \Cref{algo:incExpandCluster} togetherly describes the processing of each incremental step.\\
{\bf Re-verification of border and noise.}
The size of $\epsilon$-neighborhood of any point $v\in S_{prototype}$ may increase while querying each point in $S_{inc}$. This leads to promotion of state for vertices from  \emph{noise} $\rightarrow$ \{\emph{border, core}\} or \emph{border}  $\rightarrow$ \emph{core}. The \Cref{algo:updateNB}  describes the  process of verification for border and noise points. 

\begin{algorithm}[htb!] 
	\caption{\emph{\small incExpandCluster}}
	\label{algo:incExpandCluster}
	\small
	{\bfseries Input : }{Point: $p$, Processed sample: $S_\text{prototype}$, Unprocessed sample: $S_\text{inc}$; New Cluster Id: $\hat{k}$;\\
		Parameters for incremental DBSCAN: $\epsilon, MinPts$; 
		Lists: $LOS$, $LOC$, $LOB$, $LON$ }\\
	{\bfseries Output : }{Cluster Labels : $\mathbb{C}$, List : $\emph{LOS}$, Next Cluster Id: $\hat{k}$}
	
	\begin{algorithmic}[1]
		\State $N_{\epsilon} \leftarrow$ \emph{RangeQuery($p,\epsilon,S_\text{prototype}\cup S_\text{inc}$ )}
		\If{$|N_{\epsilon}(p)| < MinPts$}
		 
		 \State Determine whether $p$ is a noise or border
		 and update $state(p)$ and $label(p)$ accordingly
		\State \Return  $\hat{k}$
		\EndIf

		\State state(p)$\leftarrow$``core", $LOC = LOC \cup p$, $label(p) \leftarrow \hat{k}$
		\State $seeds=[]$
		\For{ $x \in N_{\epsilon}(p)$}
			\If{$x$ is not visited}
				\State $seed \leftarrow seed \cup x$
			\Else 
				\State  {RNB} (LON, LOS, $\eta$, $N_{\epsilon}$, $x$)  \Comment{Refer to \Cref{algo:updateNB} }
				\State $\mathcal{L}_{Merge} = \mathcal{L}_{Merge} \cup label(x)$
			\EndIf
		\EndFor
		\For{$q \in seeds$}
		\State $N_{\epsilon} \leftarrow$ \emph{RangeQuery ($q,\epsilon,S_\text{prototype}\cup S_\text{inc}$ )}
		\If{ $|N_{\epsilon}(q)|>MinPts$}
		\State $LOS(q) \leftarrow$ ``core", $LOC = LOC \cup q$, $label(q) \leftarrow \hat{k}$
		
		\For{ $x \in N_{\epsilon}(q)$}
		\If{($x$ is unprocessed) }
		\State $seed \leftarrow seed \cup x$ \Comment{$x$ was not in $seed$}
		\Else
		\State  {RNB} (LON, LOS, $\eta$, $N_{\epsilon}$, $x$)  \Comment{Refer to \Cref{algo:updateNB} }
		\State $\mathcal{L}_{Merge} = \mathcal{L}_{Merge} \cup label(x)$
		\EndIf	
		\EndFor	
		\Else
		\State $LOS(q) \leftarrow $ ``border", $LOB = LOB \cup q$,  $label(q) \leftarrow \hat{k}$
		\EndIf
		\EndFor
		\State $\mathcal{L}_{Merge} = \mathcal{L}_{Merge} \cup \hat{k}$
		\If{$|\mathcal{L}_{Merge}|>1$}
		\State Merge all clusters in $\mathcal{L}_{Merge}$
		\EndIf
		\State \Return $\hat{k}+1$
	\end{algorithmic}
\end{algorithm}	

\noindent
\subsection{Selection of cluster representatives}
\begin{figure}[htb!]	
	\centering
	\begin{multicols}{2}
		\subcaptionbox{Toy dataset\label{fig:toy_dataset}}{\includegraphics[width=0.5\textwidth]{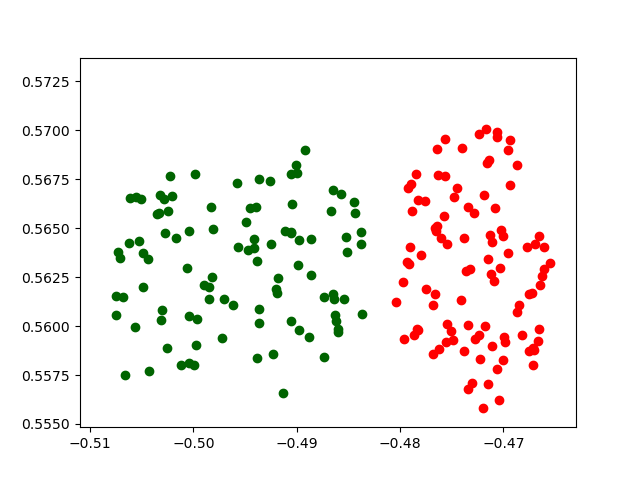}}

		\subcaptionbox{K-means partition\label{fig:toy_kmeans}}{\includegraphics[width=0.5\textwidth]{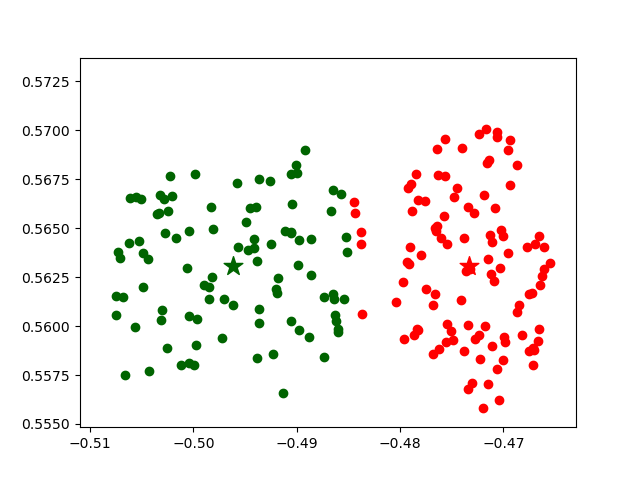}}

	\end{multicols}
	\caption{A toy example with two clusters. Centroids of K-means are shown in ``*" marking.  Kmeans assigns a few data points to red cluster which originally belong to green cluster.}
	\label{fig:toy}
\end{figure}

In a partition-based clustering algorithm, clusters are represented by the corresponding cluster centers. Data points are assigned to their nearest cluster centers. However, the scheme of a single representative for a cluster has a serious flaw. Border points of a cluster may get assigned to another nearby cluster. \Cref{fig:toy} represents a scenario of two clusters. $\exists x \in {C}_{g}$ such that $x$ is a border point of the green cluster ${C}_{g}$. \emph{K-means} assigns $x$ to the red cluster ${C}_{r}$ as its center is nearest compared to the center of ${C}_{g}$. The immediate solution is to choose multiple representatives for each cluster. The problem gets resolved to a great extent when representatives belong to the border area.  Hence, we aim to select core points that depict the contour of the arbitrary-shaped clusters.  In density-based clustering, core points are the indicator of the high-density region of the cluster. Therefore, we consider core points as the representative of the respective cluster.  Since our algorithm develops by processing a fraction of the dataset, the total number of core points is limited. 
According to our design of algorithm, test sample $S_{test}$ may be relabeled many times. If the number of representatives is equal to the number of core points, then that quantity could be a bottleneck in the sample space for a large dataset. Therefore, we need to efficiently select very few representatives from the set of core points such that the shape of each cluster could be well predicted by the set of representatives. This in turn reduces the execution time for labeling remaining unprocessed points. Hence,  we need to consider the impact of the size of representatives in labeling test sample $S_{test}$ for a very large-scale dataset where $n$ is counted in millions. Hence, we define a heuristic in \Cref{def:heuristic_R} to sieve a limited number of representatives.  {The first condition of $\Phi(q)$  selects a few core points which are nearer to the border points as representative.} The second condition of $\Phi(q)$  determines the  centroid. \Cref{def:Representative} defines the criterion of a representative.

\begin{definition}\label{def:heuristic_R}
	
	Let $\varrho_q^k =|N_{\epsilon}(q)| $ such that $q \in C_k$, and $\varrho_{max}^k=max(\varrho_q^k| q\in C_k )$. Then, a ${\Phi}$ function is defined to choose  representatives for each cluster $C_k \in C$ as follows:
	\begin{equation}
		\Phi(q) =
		\left\{
		\begin{array}{ll}
			1 & \mbox{if } \frac{\varrho_q^k}{\varrho_{max}^k}\leq {\tau} \\
			1 & \mbox{if } \frac{\varrho_q^k}{\varrho_{max}^k}= 1 \\
			0 & \mbox{Otherwise,} 
		\end{array}
		\right.
	\end{equation}
	Where $\tau$ is a user defined threshold.
\end{definition}

\begin{definition} \label{def:Representative}
	We define representatives $\mathcal{R}$ for the partition $C$ that correspond to graph $H$ such that $H$ follows \Cref{def:proto_graph} and \Cref{def:heuristic_R}. \\
	\noindent
	\begin{equation}
			\begin{array}{ll}
				\forall c \in C, & R_c=\{v ~| ~v \in V^\prime~~\text{and}~~ v ~~\text{has ``core" state and }~~ \Phi(v)=1\}
			\end{array}
	\end{equation}
\end{definition}
At the end of the incremental step, unprocessed points are assigned to their nearest core points. This scheme has two advantages: 1) it reduces the computation cost of processing the whole dataset as in \emph{DBSCAN}, and 2) the accuracy of labeling border points increases as compared to any centroid-based cluster representative system.

\subsection{Stopping criterion.}\label{sec:delta} In this step, we discuss the stopping criterion of the incremental step. We sample $S_{test}$ randomly from the remaining unprocessed points with i.i.d and without repetition. We check whether derived labels of items in  $S_{test}$ vary in the consecutive incremental step. For this we measure the change of labels in the pairs of points of $S_{test}$ as discussed in \Cref{def:stability}.
\begin{definition} \label{def:stability}
	We design a stability criterion (in terms of instability) to check whether the structure of clusters  of $G^{\prime}_\text{prototype}$ resembles the structure of clusters of $G_{\text{oracle}}$. We create two sets of cluster labels $\Omega_{test}$ and $\Omega^{\prime}_{test}$ for each point in $S_{\text{test}}$. $\Omega_{test}$ and $\Omega^{\prime}_{test}$ are created using the representatives of $G_\text{prototype}$ and $G^{\prime}_\text{prototype}$ respectively. The labeling scheme uses the following rule: \\
	\noindent
	Label $ u \in S_{test}$ with $\hat{k} \in \mathbb{C}$ where $\mathbb{C}$ is a  partition such that
	\begin{equation}
		\begin{array}{lllll}
			\hat{k} &=&\underset{\text{label}(v)}{\mathrm{argmin}} & \text{dist}(u,v) & \text{where~~~} v \in \mathcal{R}
		\end{array}
	\end{equation}
	However, if there is no cluster structure in the graph, i.e, every point is a ``noise", then every point in $S_{test}$ is labeled with $-1$.
	The $\mathbb{C}$ represents partition of either $G_\text{prototype}$ or $G^{\prime}_\text{prototype}$ and $\mathcal{R}$ represents the set of cluster representatives.\\
\noindent
	Instability  measure $\varDelta$  is defined using following function:
	\begin{equation}
		\label{eq:cv_s}
			\varDelta=\frac{1}{\binom{|S_{\text{test}}|}{2}}\sum_{1\leq i < j \leq |S_{test}|} V_{ij}(\Omega_{test}, \Omega^{\prime}_{test})
	\end{equation}
	where $V_{ij}$ measures the instability in $S_{test}$	
	\begin{equation}
		\label{eq:V_ij}
			\begin{array}{l}
				V_{ij}(\Omega_{test}, \Omega^{\prime}_{test}) = I(I(\Omega_{test}(x_{i})= \Omega_{test}(x_{j})) + I(\Omega^{\prime}_{test}(x_{i})$=$\Omega^{\prime}_{test}(x_{j}))=1)\\
			\end{array}
	\end{equation}
	where   $I(.)$ is an indicator function.
		
	\noindent
	If $x_{i}$ and $x_{j}$ gets same cluster label in $\Omega_{test}$ and  different cluster label in $\Omega^{\prime}_{test}$ (where $x_{i}, x_{j} \in S_{test}$),  or vice versa, then $I(\Omega_{test}(x_{i})= \Omega_{test}(x_{j})) + I(\Omega^{\prime}_{test}(x_{i})$=$\Omega^{\prime}_{test}(x_{j}))=1$. It is a \emph{XOR} operation. $V_{ij}$ enumerates number of pairs in $S_{test}$ which are labeled differently in two consecutive iterations.
	
\end{definition}
\begin{lemma}\label{lemma:converge}
	The algorithm converges when no new cluster emerges.
\end{lemma}
\begin{proof}
	{Since we add new samples to the prototype in every iteration, it may reveal new cluster. However, there are two plausible cases. This new cluster may be a part of the existing cluster but it is not density reachable from them at that stage. Again, this cluster may reveal truly a new cluster of the original dataset. However, this condition will stop within finite iteration since number of clusters are  constant for the dataset.
	If new cluster is part of the existing cluster and the condition of density reachability hold for the original dataset with the given $\epsilon$ and $MinPts$, then they will be eventually merged within finite iteration. Otherwise, they remain as separate cluster. Addition of new samples can not 
	merge them and eventually they become stable.\\
	In the extreme case, merging may lead to a single cluster. Therefore, the instability measure $\Delta$ will be zero. Otherwise, cluster stability is achieved, and addition of new points can not bring new information to the prototype. Hence, $\Delta$ becomes zero and iteration stops.}~~~~~~~~$\square$	
\end{proof}
\noindent
{It is assumed that the cluster structure of $G^{\prime}_\text{prototype}$ replicates the cluster structure of $X$ when the notion of stability is achieved. However, the success of this step depends on the size of $|S_{inc}|=\beta$, and the selected items in $S_\text{test}$. If $\beta$ is very small, then algorithm may converge even if all the clusters have not evolved. However, such a situation ends in having many points as noise if initial size of prototype is also very low.}
{New samples fail to provide variation in the prototype. Hence, $\Delta$ reaches to zero. Hence, we have used the principle used in} \cite{CNAK_2021} for {estimation of sample size to determine $\beta$.\\
Again, it is important for $S_{test}$ to have points from every cluster which is the best fit to $X$. As we are aiming to deal with large datasets, random sampling is the easy solution to create $S_{test}$. Therefore, the value of $\alpha=|S_{test}|$ is crucial to achieving the requirement.}

\noindent
Following terms are used in estimating the value of $\alpha$ in the subsequent lemmas and corollaries.\\
\noindent
$k$ : Number of clusters.  \\
$n$ : Total number of points. \\
$t$ : fraction of test samples. \\
$p_i$: Prob. of the $i$th cluster, $i=1,2,\ldots,k$. 
\begin{lemma}
	Probability that there  exist at least two samples of the $i$th  cluster in the test samples is given by 
	$(n t-1)^2p_i^2 t^2$.
	
\end{lemma}

\begin{proof}
	Prob. that a sample of $i$ cluster in the test set: $p_i t$.
	Number of test samples ($\alpha$): $n t$.
	Prob. that there  exist any two or more samples of the $i$ cluster  in the test data set:
	\[ 1-(1-p_i t)^{n t} - \binom {n t} {1} (p_i t) (1-p_i t)^ {(n t -1) } \]
	\[= 1- (1- p_i t)^{(nt -1) } (1 + (n t -1) p_i t) \]
	\[ \approx 1 - (1- (n t -1) P_i t) (1 + (n t -1) p_i t) \]
	\[ = 1 - ( 1 - (n t -1 ) ^2 p_i^2 t^2 \]
	\[ = ( n t -1 )^2 p_i^2 t^2  \] 
	
	\hfill{ Q.E.D}
\end{proof}

\begin{lemma}
	Probability that for every cluster there exist two or more samples is $(n^{\prime} t -1)^{2k} (\Pi_{i=1}^{k}  p_i^2) t^{2k}$.
\end{lemma}

\begin{proof}
	Probability that there  exist at least two samples of the $i$th  cluster in the test samples is given by 
	$(n^{\prime} t-1)^2p_i^2 t^2$.\\
	Hence, 	the probability that for every cluster there exist two or more samples is
	$\prod_{i=1}^{k} ((n^{\prime} t-1)^2p_i^2 t^2)=(n^{\prime} t -1)^{2k} (\Pi_{i=1}^k  p_i^2) t^{2k}$	~~~~~~~~$\square$
\end{proof}

\begin{corollary}
	The number of test samples for ensuring that there exist at least two samples of each cluster with probability $P$ is obtained from:
	\[ (n^{\prime} t -1)^{2k} (\Pi_{i=1}^k  p_i^2) t^{2k} = P \]
\end{corollary}	
One may put $P=1$, for the theoretical minimum number. Practically  $P$ can be kept high.
\begin{corollary}
	\label{cor:size}	
	For uniform probability distribution of clusters $p_i = \frac{1}{k}$. Hence, the theoretical minimum number is given by the solution of the following equation:
	\begin{equation}
		\begin{array}{l}
			n^{\prime} t^2 - t -k =0\\
			\Longrightarrow t= \frac{1}{2} (\frac{1}{n^{\prime}} \pm \sqrt{\frac{1}{{n^\prime}^2}+ \frac{4k}{n^{\prime}}})\\
			\approx \frac{1}{2n^{\prime}}+ \sqrt{\frac{k}{n^{\prime}}}
		\end{array}
		\label{eq:test_size}
	\end{equation}
\end{corollary}	
\noindent
This step aims to provide a measure of the `goodness' of the prototype. This strategy verifies whether the prototype can portray the cluster structure of the original dataset. We use the notion of ``clustering stability" to measure the `goodness' of the prototype. The notion of stability ensures that the cluster structure present in the graph replicates the cluster structure of the dataset $X$.
\subsubsection{Rectification on the size of test sample for large data}
Since the test samples are labeled  using the nearest neighbor principle, it could be a bottleneck for our algorithm when applied on very large-scale data (instances $>> 10^5$). { The size of sampled dataset is considered as $n^{\prime}$ in \Cref{cor:size}.  Similarly, $k$ also can be a bottleneck for the system. The number of clusters present in the prototype can be for computing $\alpha$. However, the value of $k$ for initial prototype will be very high. Because, small $MinPts$ is used which creates a very large number of groups. Hence, $k$ is fixed to a reasonably high value.}

\begin{theorem} \label{theorem:TimeComplexity}
	For very large data, time complexity of \emph{IPD} is independent of $n$.
\end{theorem}
\begin{proof}
	\emph{DBSCAN} requires 
	$\mathcal{O}(\gamma^2)$ for neighborhood queries.  Re-evaluation of ``core" property consumes $\mathcal{O}(\gamma )$. Each incremental step uses $\mathcal{O}((\beta)^2+\gamma  \beta)$ for neighborhood queries . It requires $\mathcal{O}(\gamma)$ for processing border and noise list.
	The termination condition requires $\mathcal{O}(\alpha^2+\alpha k)$. Hence, time complexity to identify representatives of the clusters in data space
	requires $\mathcal{O}(\gamma^2+(\gamma+(\beta^2+\gamma  \beta)+\gamma+\alpha^2+\alpha k)\times t_{ipd})$, where  $t_{ipd}$ is the total number of iterations of the incremental processing step.
	%
	Usually for large data,  $\gamma, \alpha, \beta, k <<<n$. Hence, time complexity is independent of $n$.~~~~~~~~$\square$
\end{proof}
\FloatBarrier	
%
%
%
%
%
%
%
%
%
%
%
%

\section{Experiments}\label{sec:results}
We conduct several experiments to evaluate the effectiveness and  advantage of the proposed method. We validate our algorithm by performing experiments on synthetic datasets. This helps to understand the aim of our work.  
The experiments are carried out on a workstation with 128G RAM Centos 64 bit OS, and Python 2.7 programming environment. 
We have used \emph{Euclidean} distance to measure the similarity between two points in every comparing methods since it is frequently used in the literature.
We compare results using normalized mutual information (NMI)~\cite{NMI}.
\subsection{Datasets}
\begin{figure*}[htb!]	
	\centering
	\begin{multicols}{4}

		\subcaptionbox{Aggregation\label{fig:aggregation}}{\includegraphics[width=0.25\textwidth]{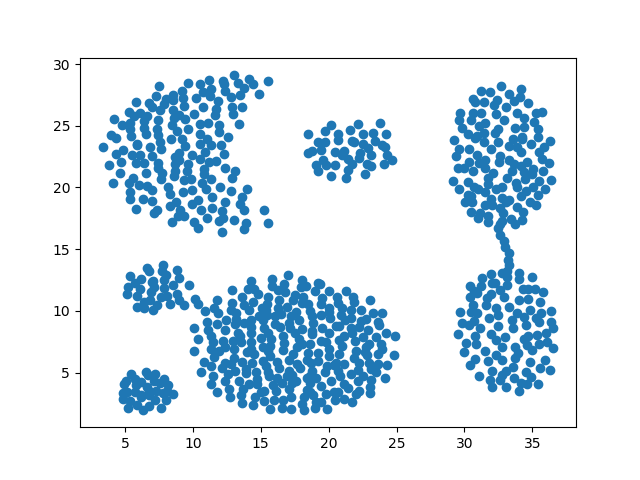}}
		
		\subcaptionbox{Compound\label{fig:Compound}}{\includegraphics[width=0.25\textwidth]{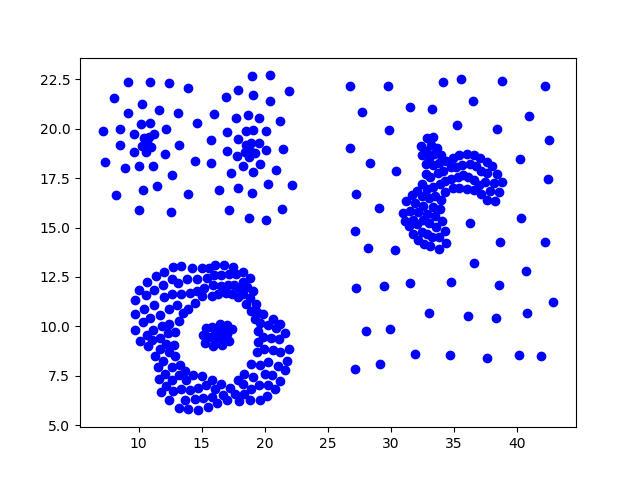}}
		
		\subcaptionbox{K30\label{fig:K30}}{\includegraphics[width=0.25\textwidth]{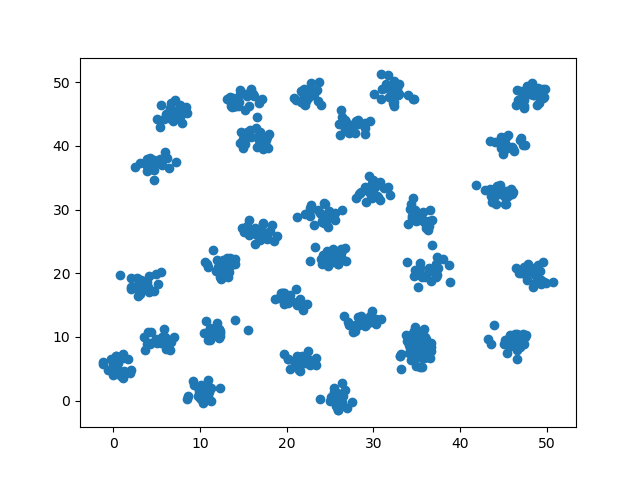}}
		
		\subcaptionbox{D31\label{fig:D31}}{\includegraphics[width=0.25\textwidth]{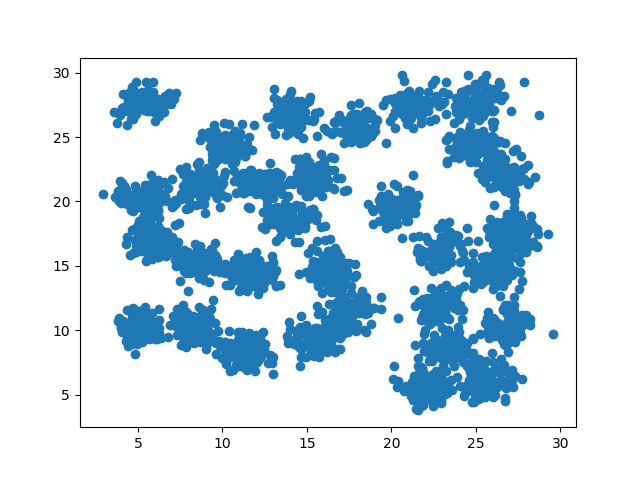}}
	\end{multicols}
		
		\begin{multicols}{4}
		
		\subcaptionbox{t4\label{fig:t4}}{\includegraphics[width=0.25\textwidth]{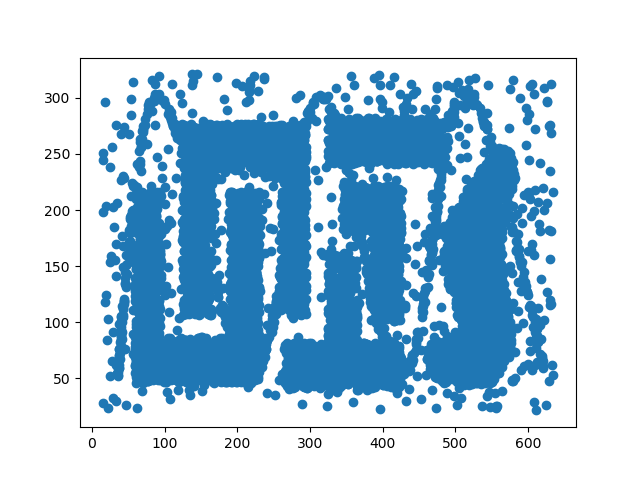}}
		
		\subcaptionbox{Aquanimals\label{fig:aquanimals}}{\includegraphics[width=0.25\textwidth]{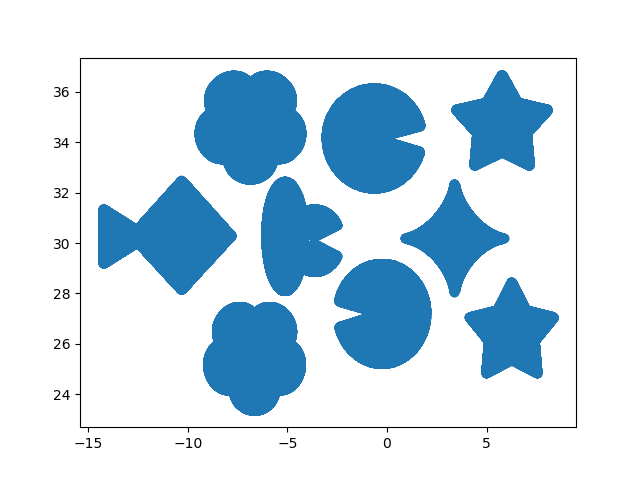}}
		
		\subcaptionbox{Artificial\label{fig:syn_flow}}{\includegraphics[width=0.25\textwidth]{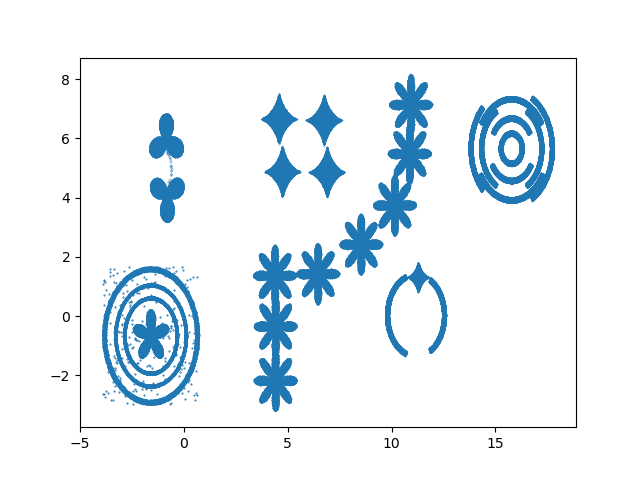}}
		
	\end{multicols}

	\caption{Scatter plot of synthetic simulations.}
	\label{fig:synthetic_simultions}
\end{figure*}

We have chosen dataset with the purpose of analyzing our algorithm on the basis of quality of clusters and capability of handling large-scale  data.  
We consider a few synthetic simulations~\cite{Aggregation} to test the applicability of our method for detecting arbitrary shaped clusters. They are widely used for cluster analysis in several research papers. Additionally, we have created two synthetic datasets that can be treated as large-scale data to check the scalability of our method.  {The description of synthetic datasets is presented in} \Cref{tab:dataset}. We have also checked its performance on a few real world datasets\\ 
\noindent
{\bf Aggregation.} Aggregation~\cite{Aggregation} is a synthetic dataset having $n=788$, $d=2$ and $K=7$. \\
\noindent
{\bf Compound.} Compound~\cite{Compound} is a synthetic dataset having $n=399$, $d=2$ and $K=6$. \\
{\bf K30.} Thirty clusters in $R^2$~\cite{CNAK_2021}. Each cluster contains 25 observations. They are an independent bivariate normal random variable with identity covariance matrix and an appropriate mean vector.  The range of each feature variable is between 0 to 100. Each mean is randomly generated. The Euclidean distance between the mean of two clusters is less than 10.\\
\noindent
{\bf D31.} D31~\cite{D31} is a synthetic dataset having $n=3100$, $d=2$ and $K=31$. Here, clusters are very closely spaced compared to $K30$.\\
\noindent
{\bf t4.8k} It~\cite{t4} is a synthetic dataset having $n=8000$, $d=2$. \\
{\bf Aquanimal.} We have created nine shaped clusters in two dimension. Each of them contains approx. $50000$ data points.\Cref{fig:aquanimals} depicts the scatter-plot of the dataset. The total number of data points are $4.65\times10^5$. We have used QGIS software to create shapes in .geojson format and fill each shape using uniform distribution in Numpy.\\
\noindent
{\bf Artificial.} We have used similar approach as aquanimal to create this dataset. We have created 29 shapes using QGIS. However, we introduced a little complexity in the dataset. For example, we have created concentric half circles and we keep their distance small. Also, we add overlapping between two clusters in flower like shapes as shown in \Cref{fig:syn_flow}. We add noise in the set of concentric circles. We add connection between two distant clusters which are having flower like shape in the \Cref{fig:syn_flow}.\\ 
{\bf PAMAP2} is a physical activity monitoring dataset having n=1,921,431 and d=39. \\
{\bf MNIST} is a digit dataset having n= 70000 and d=784.\\ 

\begin{algorithm}[htb!] 
	\caption{\emph{\small Incremental Prototype based \emph{DBSCAN} (IPD)}}
	\label{algo:IPD}
	\small
	{\bfseries Input : }{Dataset: $X$;DBSCAN parameters : $\epsilon, ~~\text{Minpts}$; $|S_{prototype}|$: $\gamma$;~~$|S_{inc}|$ :$\beta$;  threshold  : $\tau$\\
	}
	{\bfseries Output : }{ Cluster Labels : $\mathbb{C}$, Representatives : $\mathcal{R}$ }
	
	\begin{algorithmic}[1]
		\State $LOS=$\{``unknown"\}$_{i=1}^{n}$, $N_{\epsilon}=\{\{\}\}_{i=1}^n$
		
		\State $LOC=\{\}, LOB=\{\}, LON=\{\}$, $\varDelta=\infty$
		\State $S_{\text{prototype}} \leftarrow$ Sample $\gamma$ points randomly from $X$ ,  $X_{rem} \leftarrow X-S_{\text{prototype}}$ 
		
		\State $\eta \leftarrow$  {EIM}($S_{\text{prototype}}, \epsilon, MinPts, \eta=1$) \Comment{\Cref{algo:EIM}}

		\State $nextId \leftarrow$\emph{DBSCAN}$ (S_{\text{prototype}}, \epsilon, \eta)$\Comment{ $N_{\epsilon}, state, core, border, noise$ is updated by \emph{DBSCAN}}

		\State  $C \leftarrow$ labels of $S_{\text{prototype}}$ assigned by \emph{DBSCAN}.
		\State $\alpha \leftarrow$ \emph{computeTestSize}($C, n$) \Comment{Follow Lemma 4.3}
		
		\State $S_{\text{test}} \leftarrow$ Sample $\alpha$ points randomly from $X_{rem}$; $X_{rem} \leftarrow X_{rem}-S_{\text{test}}$.

		\State  $\Omega_\text{test} \leftarrow$ Obtain Cluster labels for $S_\text{test}$ as described in \Cref{def:sinc}.
		
		\While{$\varDelta > 0$ and $|X_{rem}|>0$}
		\State $S_\text{inc} \leftarrow$ Sample $\beta$ points randomly from $X_\text{rem}$; $X_\text{rem} \leftarrow X_\text{rem}-S_\text{inc}$

		\If{$\eta < MinPts$}
		\State $\eta \leftarrow$  EIM($S_{\text{prototype}}, \epsilon, MinPts, \eta$) \Comment{\Cref{algo:EIM}}
		\State  {RC}(LOC, LOS, $\eta$, $N_\epsilon$) \Comment{\Cref{algo:updateCore}}
		\EndIf

		\State $nextId\leftarrow$\emph{incDBSCAN}($S_\text{inc}, \epsilon, \eta$, $nextId$) \Comment{\Cref{algo:incDBSCAN}}

		\State $\Omega^{\prime}_\text{test} \leftarrow$ Obtain Cluster labels for $S_\text{test}$ as described in \Cref{def:sinc}.
		\State $\Delta \leftarrow$ \emph{computeStability}($\Omega^{\prime}_\text{test}, \Omega_\text{test}$) \Comment{\Cref{def:sinc}}
		\State $\alpha^{\prime} \leftarrow$ \emph{computeTestSize}($C, n$) \Comment{Follow Lemma 4.3}
		\If{$\alpha^{\prime}-\alpha>0$ and $\alpha^{\prime}-\alpha>|X_{rem}|$} 
		\State $S_{test}^{new} \leftarrow$ Sample ($\alpha^{\prime}-\alpha$) points from $X_{rem}$
		\State $S_\text{test} \leftarrow S_\text{test} \cup$ $S_{test}^{new}$; $X_{rem}\leftarrow X_{rem}-S_{test}^{new}$
		
		\State Compute $\mathcal{R}$ from $C^\prime$  as defined in \Cref{def:Representative}.
		
		\State  $\Omega_\text{test} \leftarrow$ Obtain Cluster labels for $S_\text{test}$ as described in \Cref{def:sinc}.
		\EndIf
		\State $S_{\text{prototype}} \leftarrow S_{\text{prototype}} \cup S_{\text{inc}}$
		\State $C^{\prime} \leftarrow$  labels of $S_{\text{prototype}}$ assigned by \emph{incDBSCAN}  
		\If{$\varDelta == 0$ and $\eta<MinPts$}
		\State $\varDelta \leftarrow 1$,  $\eta=MinPts$
		
		\State   RC(LOC, LOS, $\eta$, $N_\epsilon$) \Comment{\Cref{algo:updateCore}}
		\EndIf
		\EndWhile
		\State Compute $\mathcal{R}$ from $C^\prime$  as defined in \Cref{def:Representative}.
		
		\State $\mathcal{C} \leftarrow $ Each point of  $X_{\text{rem}}$ is labeled with the label of nearest representative $r \in \mathcal{R}$.
		\State $\mathbb{C} \leftarrow \mathcal{C} \cup C^{\prime}$	
	\end{algorithmic}
\end{algorithm}
\noindent

\begin{table}[htb!]
	\caption{Description of datasets.}
	\resizebox{\columnwidth}{!}{
		\makebox[\linewidth]{
			\begin{tabular}{cccc}
				\hline
				{Dataset} & $\#$elements & $\#$features&K 
				\\ \hline
				Aggregation & 788 & 2&7 \\
				Compound & 399 & 2& 5 \\
				K30 & 750 & 2&30 \\
				D31 & 3100 & 2&31 \\
				t4.8k & 8000 & 2&- \\
				Artificial & 320050 & 2 & 41        \\
				Aquanimals & 4650000 & 2&9 \\
				\hline
			\end{tabular}
		}
	}
	\label{tab:dataset}
\end{table}

\subsection{An example of cluster procedure}
\begin{figure*}[htb!]	
	\centering
	\begin{multicols}{4}
		\subcaptionbox{DBSCAN\label{fig:agg1}}{\includegraphics[width=0.27\textwidth]{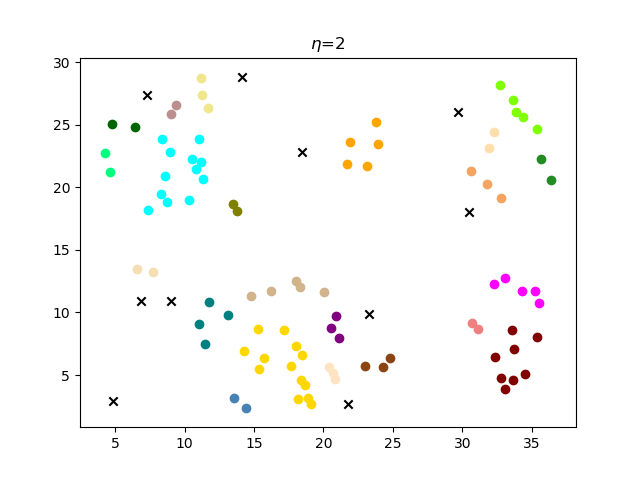}}
		
		\subcaptionbox{iteration=1\label{fig:agg2}}{\includegraphics[width=0.27\textwidth]{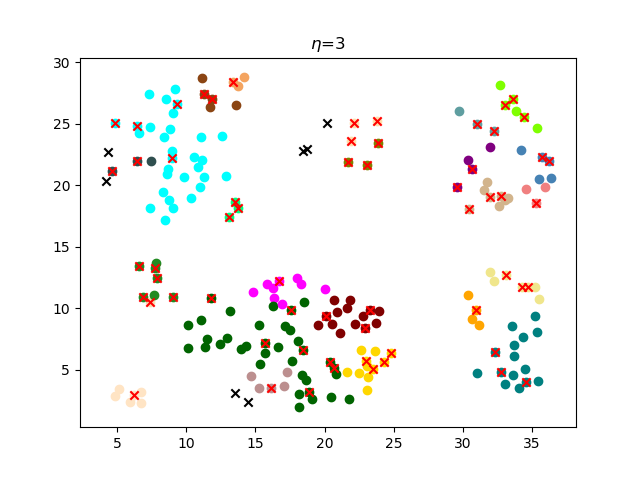}}

		\subcaptionbox{iteration=2\label{fig:agg3}}{\includegraphics[width=0.27\textwidth]{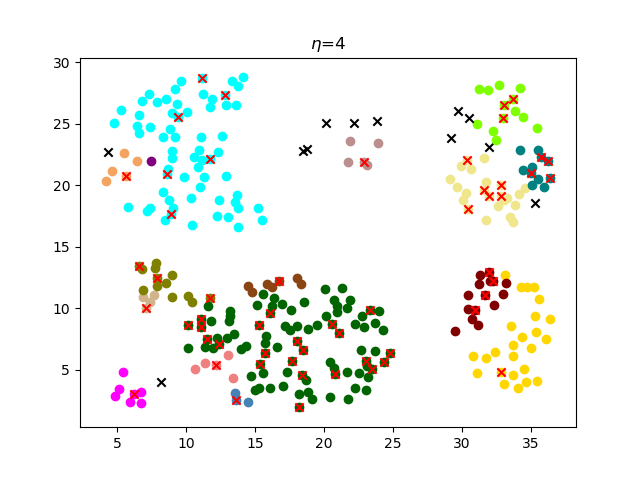}}
		
		\subcaptionbox{iteration=3 \label{fig:agg4}}{\includegraphics[width=0.27\textwidth]{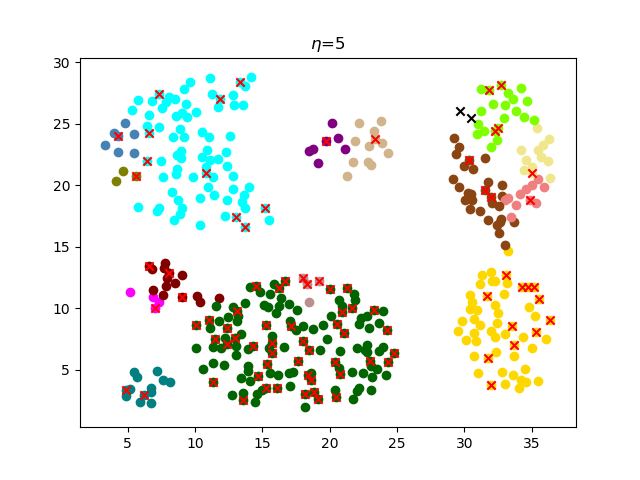}}
		
	\end{multicols}
	
	\begin{multicols}{4}
		\subcaptionbox{iteration=4\label{fig:agg5}}{\includegraphics[width=0.27\textwidth]{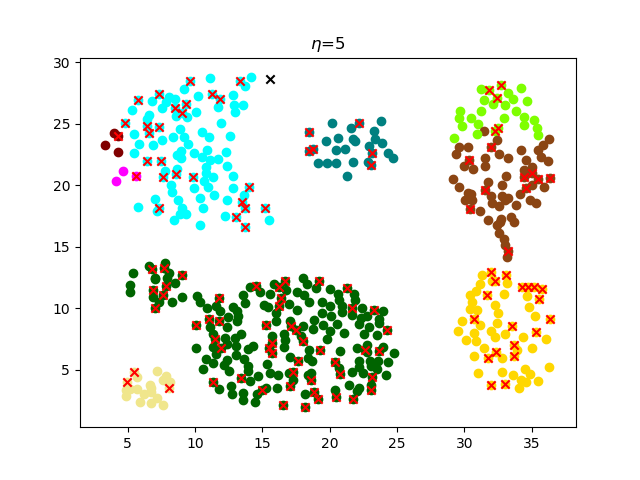}}
		
		\subcaptionbox{iteration=5\label{fig:agg6}}{\includegraphics[width=0.27\textwidth]{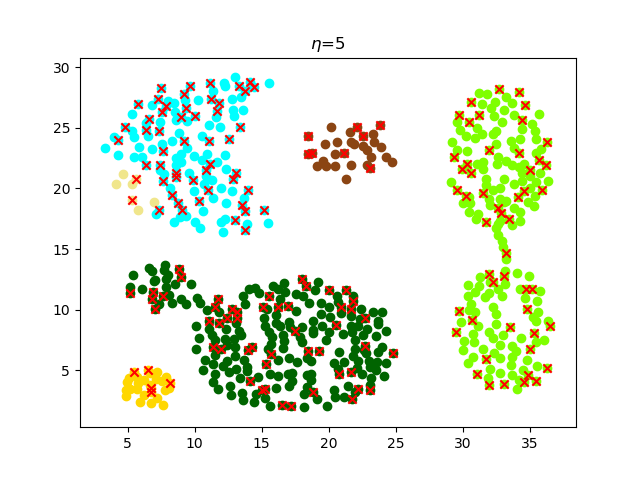}}

		\subcaptionbox{iteration=6\label{fig:agg7}}{\includegraphics[width=0.27\textwidth]{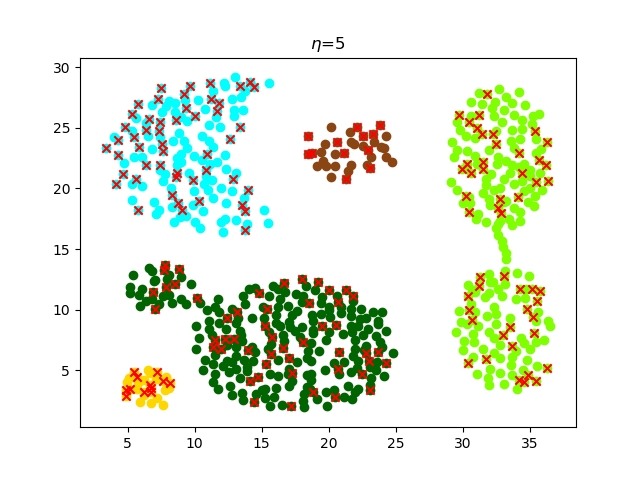}}
		
		\subcaptionbox{final clustering \label{fig:agg8}}{\includegraphics[width=0.27\textwidth]{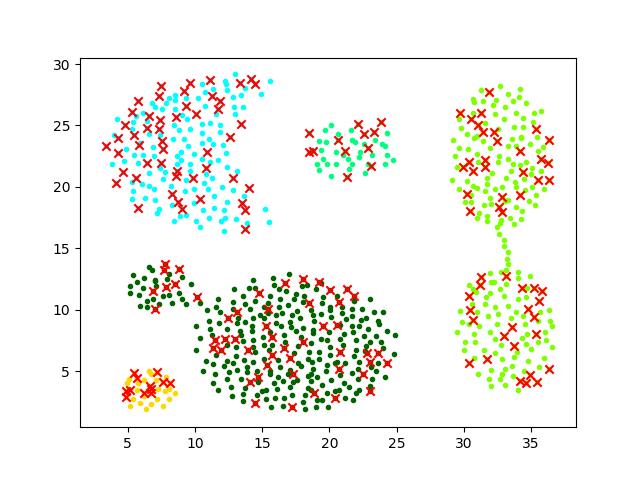}}
		
	\end{multicols}
	
	\caption{The main clustering procedures by using {\emph IPD} with $\epsilon = 2$, $MinPts = 5$, $\gamma=12\%$ of the dataset, $\beta=12\%$ of the dataset, and $\tau=0.5$. (a) shows result of \emph{DBSCAN}  with $\eta=2$. (b)-(g) show incremental processing. (h) shows final clustering result of whole dataset.  Representatives are shown in red color with marker 'X'. Noises are shown in black color with marker 'X'.}
	\label{fig:example_IPD}
\end{figure*}

We present an example here to illustrate the whole procedure
of our approach on a 2-dimensional synthetic data (\Cref{fig:aggregation} ) with
$\epsilon =2$, $MinPts = 5$ . \Cref{fig:example_IPD} shows \emph{DBSCAN} and incremental steps  of {\emph IPD}. \Cref{fig:agg6} is the clustering outcome for whole dataset.

\begin{figure*}[htb!]	
	\centering
	\begin{multicols}{5}

		\subcaptionbox{ $1.000, 0.894$\label{fig:k5_dbscan}}{\includegraphics[width=0.2\textwidth]{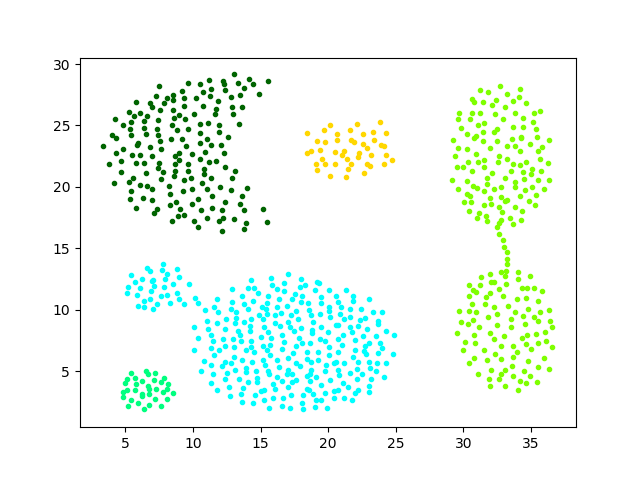}}

		\subcaptionbox{$0.950, 0.932$\label{fig:k5}}{\includegraphics[width=0.2\textwidth]{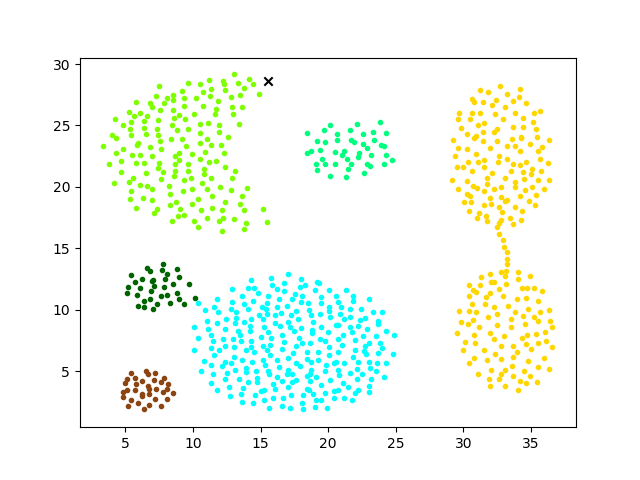}}
		
		\subcaptionbox{$0.933, 0.954$\label{fig:k6_1}}{\includegraphics[width=0.2\textwidth]{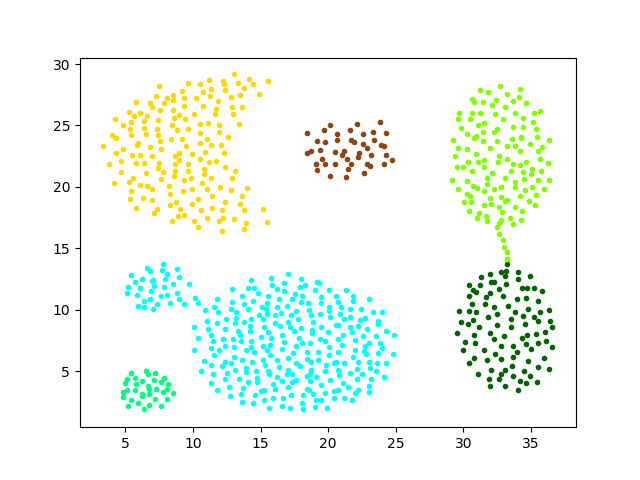}}
		
		\subcaptionbox{$0.920, 0.939$\label{fig:k6_2}}{\includegraphics[width=0.2\textwidth]{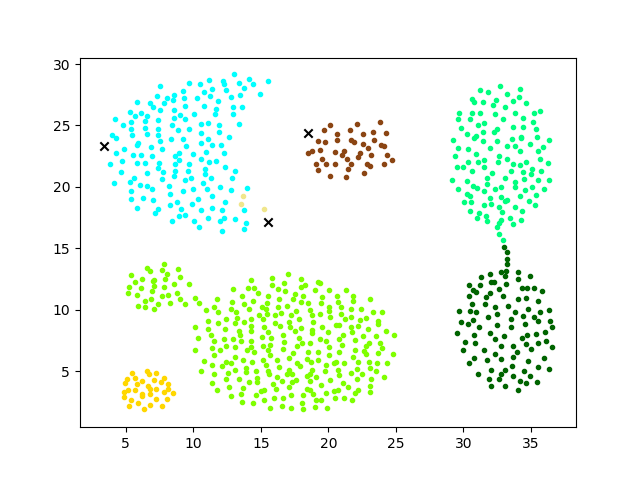}}

		\subcaptionbox{$0.886, 0.997$\label{fig:k7}}{\includegraphics[width=0.2\textwidth]{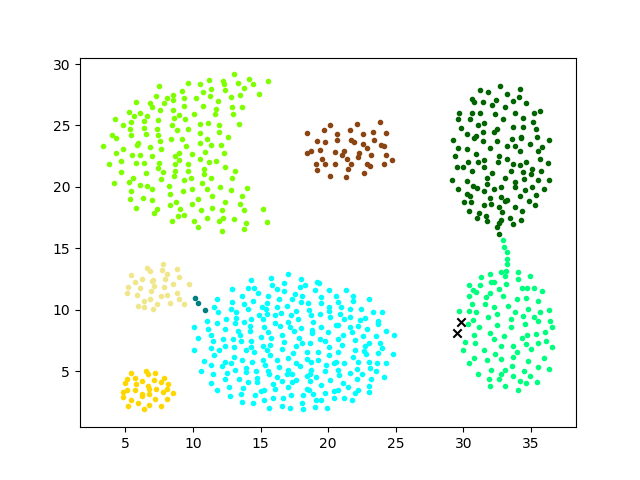}}

	\end{multicols}	
	\caption{Several clustering structure of \emph{Aggregation} dataset.  (a)-(e):prediction by \emph{IPD}. Parameters used $\epsilon=2$, $MinPt=2$, $\gamma=12\%$ of the dataset, $\beta=12\%$ of the dataset, and $\tau=0.5$. $(nmi_{dbscan}, nmi_{gt})$ are reported in the subcaption where $nmi_{dbscan}$ compares prediction of IPD and prediction of DBSCAN for the same $\epsilon$ and $MinPts$, and $nmi_{dbscan}$ compares prediction of IPD with the ground truth}
	\label{fig:randomness}
\end{figure*}

\subsection{Effect of randomness on clustering structure}~\label{sec:randomness}
Since \emph{IPD} is a sampling-based method, random behavior could be observed. To understand the effect of randomness we have executed IPD several times with the same parameters. We have chosen \emph{Aggregation} dataset to perform such a test and we observe interesting characteristics of our algorithm. We have kept $\epsilon=2$, $MinPts=5$, $\gamma=12\%$ of the dataset, $\beta=12\%$ of the dataset, and $\tau=0.5$ for the experiment and run the algorithm for $50$ times. We depict a few clustering structure produced by \emph{IPD} in $50$ such executions in \Cref{fig:randomness}. We observe that our algorithm produces  $K=5$, $K=6$ and  $K=7$ for $16\%$, $42\%$ and $12\%$ times respectively.  We have shown normalized mutual information between prediction of \emph{IPD} and \emph{DBSCAN} and prediction of \emph{IPD} and the ground truth  in  \Cref{fig:randomness}. This phenomenon suggests that our method is effective in capturing multiple suitable clustering structures.

\begin{figure*}[htb!]	
	
	\begin{multicols}{3}
		
		\subcaptionbox{Aggregation\label{fig:agg_bar} }{\includegraphics[width=0.33\textwidth ]{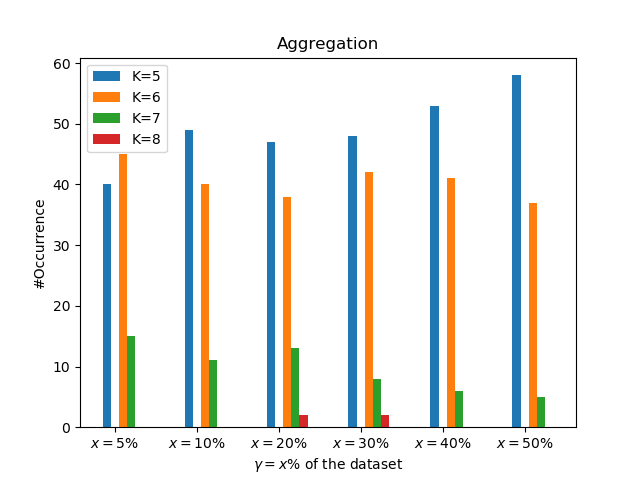}}
		
		\subcaptionbox{Compound\label{fig:comp_bar} }{\includegraphics[width=0.33\textwidth ]{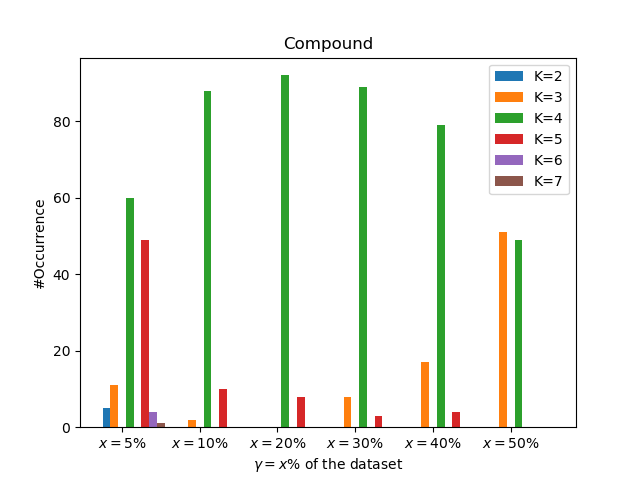}}
		
		\subcaptionbox{D31\label{fig:d31_bar} }{\includegraphics[width=0.33 \textwidth  ]{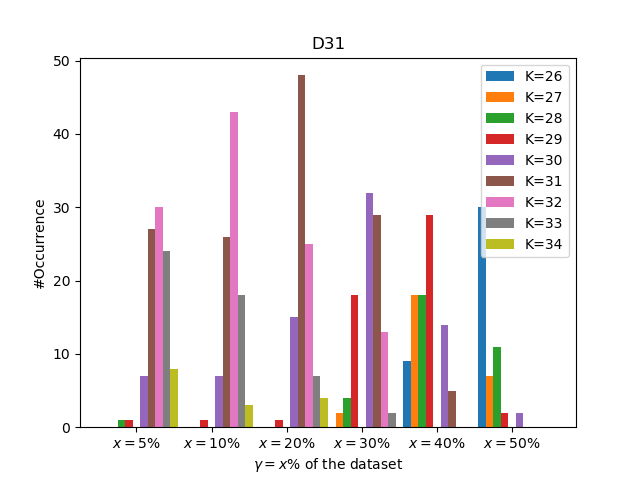}}
		
	\end{multicols}
	
	\caption{Effect of $\gamma$ on clustering result. }
	\label{fig:gamma}
\end{figure*}

\subsection{The effect of {initial prototype size} {$\gamma$} on IPD}
To study the effect of $\gamma$, we run our algorithm 100 times on each of the chosen sizes of $\gamma$. We have shown the type of cluster structure (in terms of cluster number) obtained for various values of $\gamma$ in \Cref{fig:gamma}.
We have chosen \emph{Aggregation}, \emph{Compound} and \emph{D31} datasets for this experiment. \Cref{fig:agg_bar} suggests that {estimated cluster number} $K=5$ predominates when the initial size of the prototype is large.  \emph{DBSCAN} also provides a similar cluster structure. However, with the small size of the initial prototype, several cluster structure is observed. We observe the similar phenomenon for \emph{Compound} dataset in \Cref{fig:comp_bar}. On the other hand  {\emph D31} shows an interesting behavior in \Cref{fig:d31_bar}. When the initial prototype size is small, the resulting cluster structure contains more than 31 clusters most of the time. But, the resulting cluster structure contains clusters between 26 to 31, when that size is large. Inter-cluster distance is small in D31. With the large size of samples, a few clusters appear as a single cluster. Hence, a small number of clusters predominates.                                                                                                                                                                                                                                                                                                                                                                                                                                                                                                                                                                                                                                                                                                                                                                                                                                                                                                                                     
We draw the following observation from this fact: Our method reveals the existence of hierarchy on the dataset on several runs with an initial prototype of small size. The clustering structure of the initial prototype is obtained by \emph{DBSCAN}. Therefore, a large size prototype tends to produce a similar cluster structure as in  \emph{DBSCAN}.

\subsection{Quality of clustering}
\begin{figure*}[htb!]	
	
	\begin{multicols}{5}
		
		\subcaptionbox{Aggregation\label{fig:agg_dbscan} }{\includegraphics[width=0.2\textwidth ]{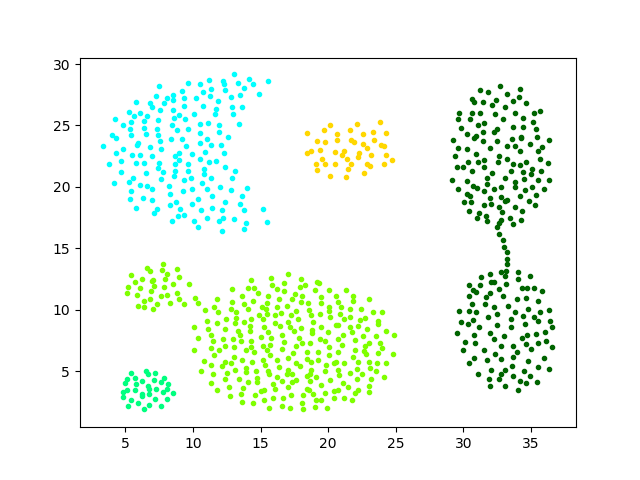}}
		
		\subcaptionbox{Compound\label{fig:compound_dbscan} }{\includegraphics[width=0.2\textwidth ]{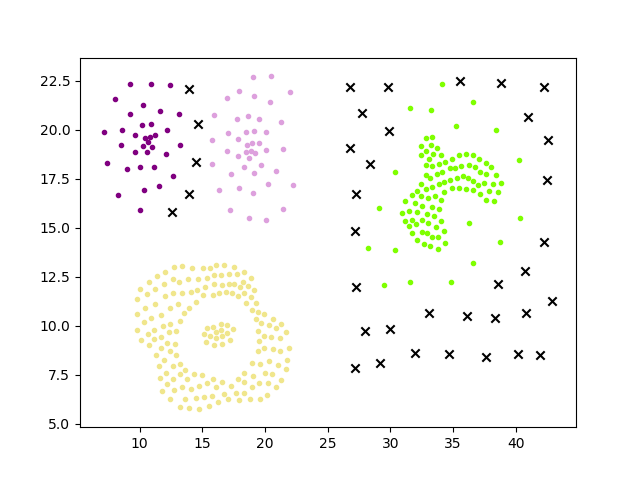}}
		
		\subcaptionbox{D31\label{fig:d31_dbscan} }{\includegraphics[width=0.2\textwidth ]{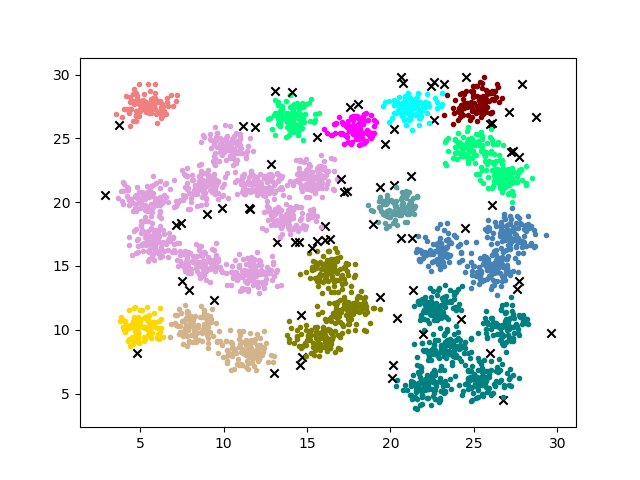}}
		
		\subcaptionbox{K30\label{fig:k30_dbscan} }{\includegraphics[width=0.2\textwidth ]{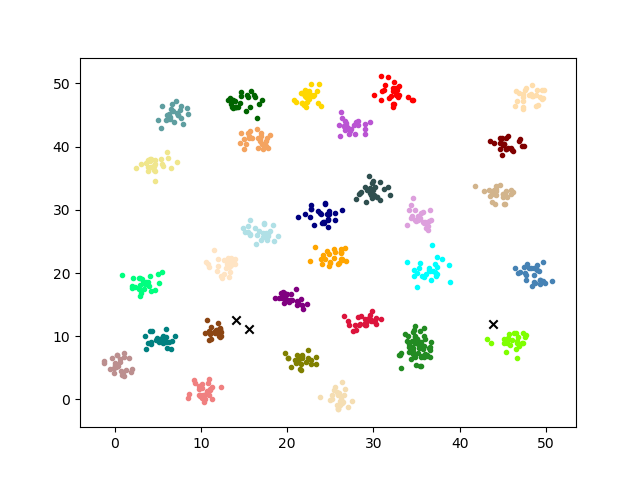}}
		
		\subcaptionbox{t4\label{fig:t4_dbscan} }{\includegraphics[width=0.2\textwidth ]{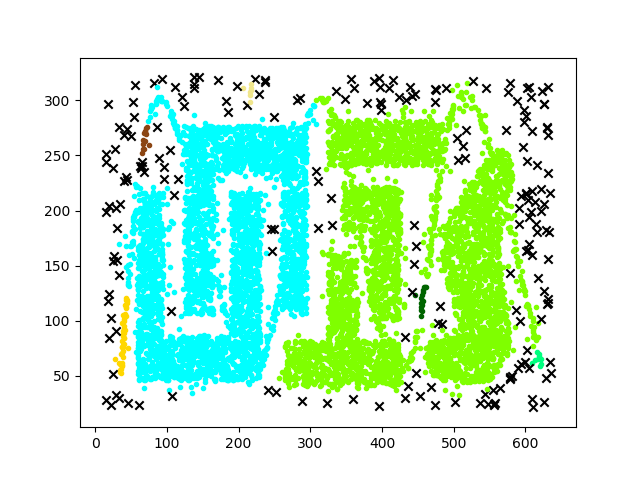}}
		
	\end{multicols}
	
	\begin{multicols}{5}

	\subcaptionbox{Aggregation\label{fig:ipd} }{\includegraphics[width=0.2\textwidth ]{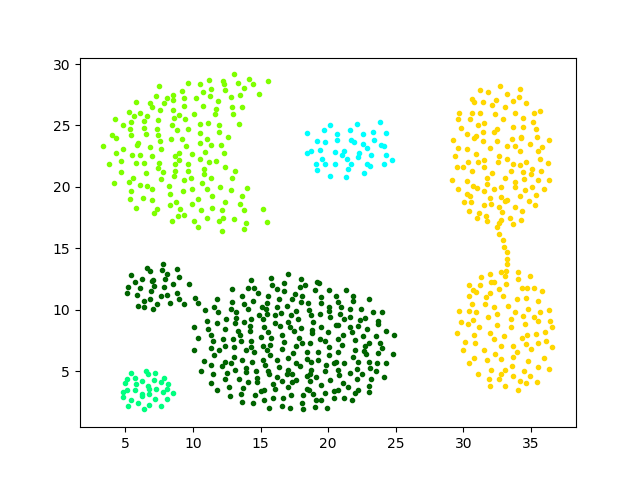}}
	
	\subcaptionbox{Compound\label{fig:compound_ipd} }{\includegraphics[width=0.2\textwidth ]{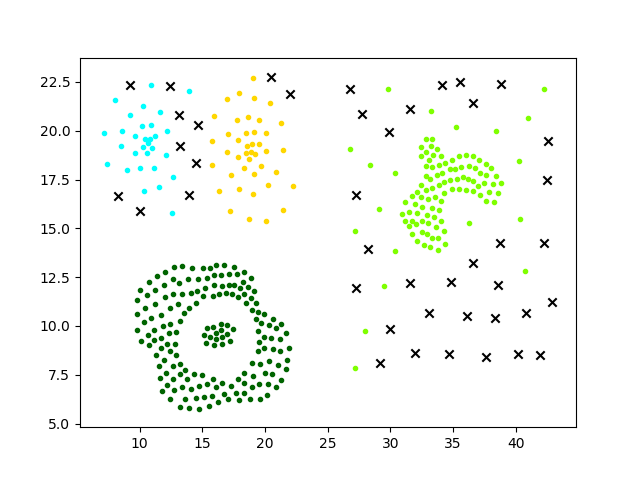}}
	
	\subcaptionbox{D31\label{fig:d31_ipd} }{\includegraphics[width=0.2\textwidth ]{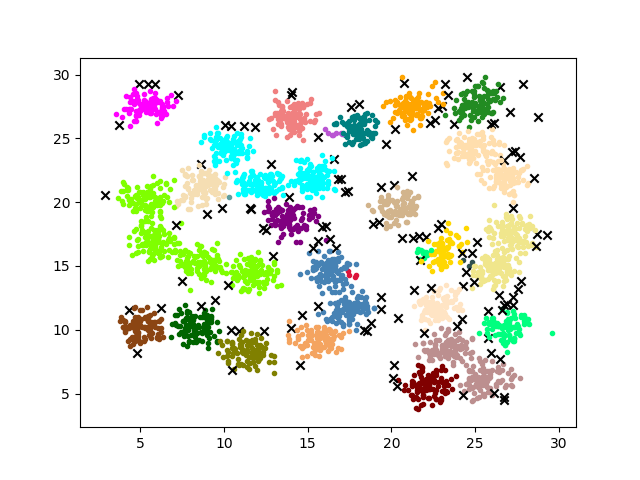}}
	
	\subcaptionbox{K30\label{fig:k30_ipd} }{\includegraphics[width=0.2\textwidth ]{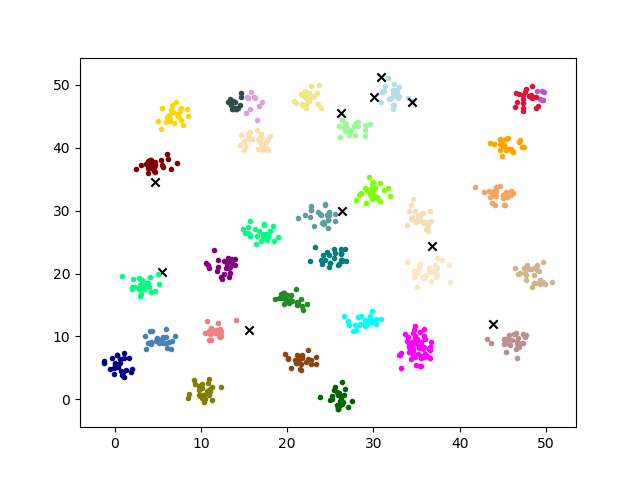}}
	
	\subcaptionbox{t4\label{fig:t4_ipd} }{\includegraphics[width=0.2\textwidth ]{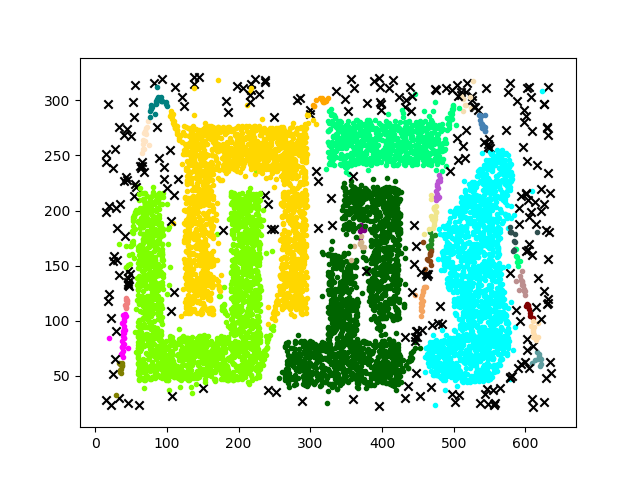}}
	\end{multicols}

	\caption{Clustering results of \emph{DBSCAN} ((a)-(e)) and \emph{IPD} ((f)-(j)) on small scale synthetic dataset }
	\label{fig:comp}
\end{figure*}

We examine quality of clustering of our method with respect to DBSCAN and the ground-truth (gt). We use normalized mutual information (NMI)~\cite{NMI} to compare two labels set $\mathcal{L}_Z,\mathcal{L}_{IPD}$ where $\mathcal{L}$ represents labels of data-points and $\mathcal{L}_Z$ represents either Labels predicted by DBSCAN or the available ground-truth. We have chosen small scale synthetic dataset to examiine the performance of IPD in identifying clustering structure. Since \emph{IPD} is a sampling based method, we run the algorithm 50 times and report mean and standard deviation of the metric. 
\noindent
\Cref{tab:result} depicts the results and \Cref{fig:comp} provide
the 2D visualization of the clustering outcome. It suggests that outcome of \emph{IPD} is not exactly similar to \emph{DBSCAN} for all the datasets. \Cref{fig:comp} suggests that IPD can capture more  detailed clustering structure compared to \emph{DBSCAN} for a given $\epsilon$ and $MinPts$. For example, \emph{IPD} captures 22 clusters, and \emph{DBSCAN} captures 12 clusters for D31. However, there are 31 clusters present in an overlapping manner. Similarly, \emph{IPD} captures detailed shaped clusters for t4 compared to DBSCAN.
Moreover, \Cref{tab:result} also suggests that several rational cluster structure could be identified by \emph{IPD} for a dataset.  \Cref{fig:randomness}  shows such incident for Aggregation dataset.
\begin{table}[htb!]
	\centering
	\caption{Performance of IPD for measuring clustering quality using small scale synthetic datasets. '*' indicates other non predominating cluster structure.}
	\resizebox{1\columnwidth}{!}{
						\makebox[\linewidth]{
	\begin{tabular}{cccclc}
		\hline
		\multirow{2}{*}{Dataset} & \multicolumn{2}{c}{Parameters} & \multicolumn{2}{c}{NMI $(\mathcal{L}_{Z},\mathcal{L}_{IPD})$} & \multirow{2}{*}{\#clusters (frequency)} \\ \cline{2-5}
		& ($\epsilon, MinPts$) & $\alpha, \beta, \tau$ & $Z=$ dbscan & $Z=$ gt &  \\\hline
		Aggregation & 2, 5 & 12\%, 12\%, 0.5 & 0.96 $\pm$ 0.04 & 0.92 $\pm$ 0.04 & 5 (16\%), 6 (42\%), 7 (24\%) \\
		Compound & 2.28, 11 & 15\%, 10\%, 0.5 & 0.84 $\pm$ 0.04 & 0.82 $\pm$ 0.02 & 3 (10\%), 4 (64\%), 5(20\%) \\
		D31 & 0.65, 5, 0.5 & 10\%, 10\%, 0.5 & 0.82 $\pm$ 0.02 & 0.86 $\pm$ 0.02 & 31(12\%), * \\
		K30 & 2.09, 6, 0.5 & 15\%, 10\%, 0.5 & 0.99 $\pm$ 0.01 & 0.98 $\pm$ 0.01 & 30 (58\%), 31 (32\%), 32(10\%) \\
		t4 & 10, 7, 0.5 & 15\%, 10\%, 0.5 & 0.67 $\pm$ 0.02 & \multicolumn{1}{c}{-} & 22 (28\%),  24 (18\%), * \\\hline
	\end{tabular}
}}
\label{tab:result}
\end{table}

\subsection{Experiments on large data}
\subsubsection{Comparable Methods} 
The basic principle of our method is similar to \emph{DBSCAN}. We use a sampling-based strategy to gain scalability. Our prime concern is to obtain stability for the cluster structure in sample space. We select representatives for the cluster structure once stability is achieved and the algorithm do not process further any unprocessed data points. Several methods~\cite{BLOCK-DBSCAN2021, IncAnyDBC2020, grid2019} have been developed to use \emph{DBSCAN} for large data. None of them examine the quality of cluster structure  for large data. They only check scalability. Their major concern is to reduce the number of query points while processing the whole dataset. Their output approximates DBSCAN output. Since the motivation of this work is quite different from the existing scalable DBSCAN algorithms, we have chosen only one of them, \emph{IncAnyDBC}~\cite{IncAnyDBC2020} for comparison.

Since our method generates cluster representatives by processing a fraction of the dataset (randomly sampled), it makes our algorithm efficient in time and storage. Hence, we compare our method with that of \emph{CNAK}, which is the sampling-based \emph{K-means} method where $K$ is automatically learned during its execution. It is also capable of handling large data. We will be able to compare the performance of a multi-representative scheme with a single representative one.
\subsubsection{Analysis of clustering for large dataset}
\emph{IPD} processes a fraction of the dataset to generate representative points for the cluster structures present in the data. Unprocessed points are labeled using the 1-NN rule.
In contrast, \emph{IncAnyDBSCAN} processes the whole dataset to determine cluster labels. \emph{IncAnyDBSCAN} and \emph{IPD} have quite different motivations. In IncAnyDBSCAN, the authors attempt to provide results similar to \emph{DBSCAN} for large datasets in reasonable time.
In contrast, \emph{IPD} attempts to identify representatives that can accurately reflect the existing clustering structure for the whole dataset. Hence, runtime for \emph{IPD} and \emph{CNAK} reflects the time needed to identify  representatives.
\begin{table}[htb!]
	\centering
	\caption{Runtime  comparison on large scale data}
	\resizebox{1\columnwidth}{!}{
		\makebox[\linewidth]{
	\begin{tabular}{c|c|lcc|cccc|ccc}
		\hline
		\multirow{2}{*}{Dataset} & \multirow{2}{*}{$\epsilon, MinPts$} & \multicolumn{3}{c|}{IncAnyDBC} & \multicolumn{4}{c|}{IPD} & \multicolumn{3}{c}{CNAK} \\ \cline{3-12} 
		&  & $\alpha_{i}, \beta_i, s$ & runtime & K & $\tau$ & runtime & K & $X_{processed} $ & $K_{max}$ & runtime & K \\ \hline
		\multirow{4}{*}{Artificial} & 0.3, 10 & \multirow{4}{*}{\begin{tabular}[c]{@{}l@{}}(256, \\ 256, \\ 100)\end{tabular}} & 64 (s) & 12 & \multirow{4}{*}{0.3} & 1567 (s) & 12 & 318468 & \multirow{4}{*}{50} & \multirow{4}{*}{126 (s)} & \multirow{4}{*}{6} \\
		& 0.4, 10 &  & 46 (s) & 9 &  & 48 (s) & 12 & 11500 &  &  &  \\
		& 0.6, 10 &  & 44 (s) & 6 &  & 26 (s) & 9 & 7500 &  &  &  \\
		& 1, 10 &  & 175 (s) & 6 &  & 19 (s) & 7 & 5500 &  &  &  \\ \hline
		\multirow{3}{*}{Aquanimal} & 0.4, 10 & \multirow{3}{*}{\begin{tabular}[c]{@{}l@{}}(256, \\ 256, \\ 100)\end{tabular}} & 3141 (s) & 9 & \multirow{3}{*}{0.3} & 33 (s) & 10 & 13950 & \multirow{3}{*}{20} & \multirow{3}{*}{424 (s)} & \multirow{3}{*}{8} \\
		& 1.0, 10 &  & 36 (hrs) & 7 &  & 28 (s) & 9 & 13950 &  &  &  \\
		& 100, 100 &  & 109 (s) & 1 &  & 17 (s) & 1 & 6975 &  &  & \\\hline
		\multirow{3}{*}{PAMAP2} & 200, 100 & \multirow{3}{*}{\begin{tabular}[c]{@{}l@{}}(256,\\ 256,\\ 100)\end{tabular}} & 129 (s) & 1 & 0.3 & 116 (s) & 1 & 28824 & \multirow{3}{*}{31} & \multirow{3}{*}{692 (s)} & \multirow{3}{*}{1} \\
		& 350, 150 &  & 63 (s) & 1 &  & 116 (s) & 1 & 28824 &  &  &  \\
		& 3500, 500 &  & 62 (s) & 1 &  & 124 (s) & 1 & 28824 &  &  &  \\ \hline
		\multirow{3}{*}{MNIST} & 700, 10 & \multirow{3}{*}{\begin{tabular}[c]{@{}l@{}}( 256, \\  256,\\  100)\end{tabular}} & 1.36 (hrs) & 3 & 0.3 & 118 (s) & 1 & 8400 & \multirow{3}{*}{20} & \multirow{3}{*}{198 (s)} & \multirow{3}{*}{1} \\
		& 1000, 10 &  & 1.29 (hrs) & 32 &  & 1.2 (hrs) & 22 & 68418 &  &  &  \\
		& 1330, 458 &  & 1.39 (hrs) & 1 &  & 26 (hrs) & 1 & 68418 &  &  &    \\\hline
	
	\end{tabular}
}}
\end{table}

\begin{figure*}[htb!]	
	
	\begin{multicols}{4}
		
		\subcaptionbox{$\epsilon=0.3, MinPts=10$\label{fig:arti_1} }{\includegraphics[width=0.25\textwidth ]{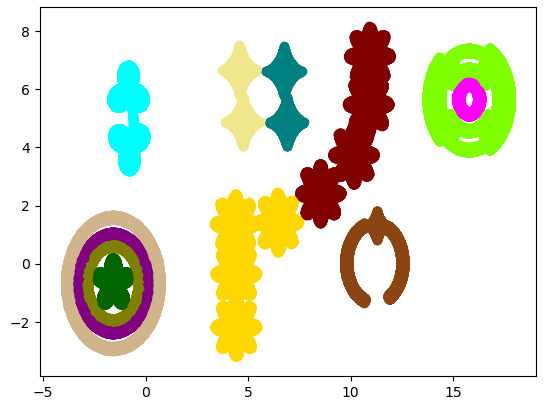}}
		
		\subcaptionbox{$\epsilon=0.4, MinPts=10$\label{fig:arti_2} }{\includegraphics[width=0.25\textwidth ]{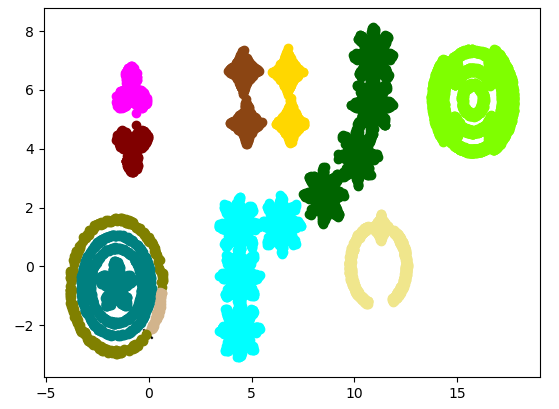}}
		
		\subcaptionbox{$\epsilon=0.6, MinPts=10$\label{fig:arti_3} }{\includegraphics[width=0.25\textwidth ]{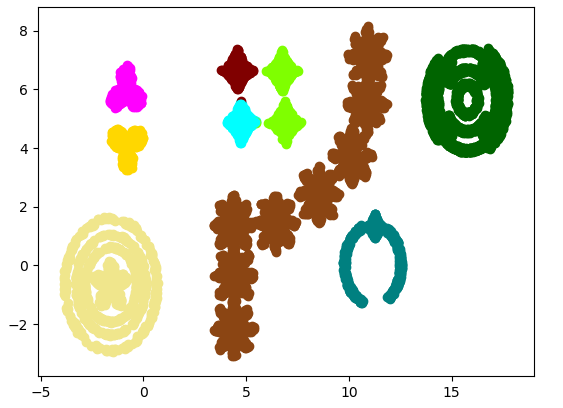}}
		
		\subcaptionbox{$\epsilon=1.0, MinPts=10$\label{fig:arti_4} }{\includegraphics[width=0.25\textwidth ]{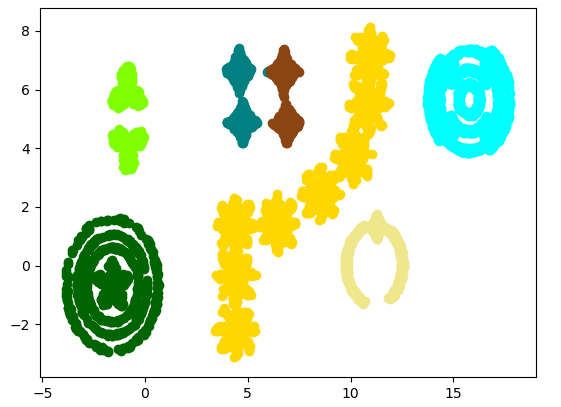}}

	\end{multicols}

	\begin{multicols}{4}
	
	\subcaptionbox{$\epsilon=0.3, MinPts=10$\label{fig:arti_11} }{\includegraphics[width=0.25\textwidth ]{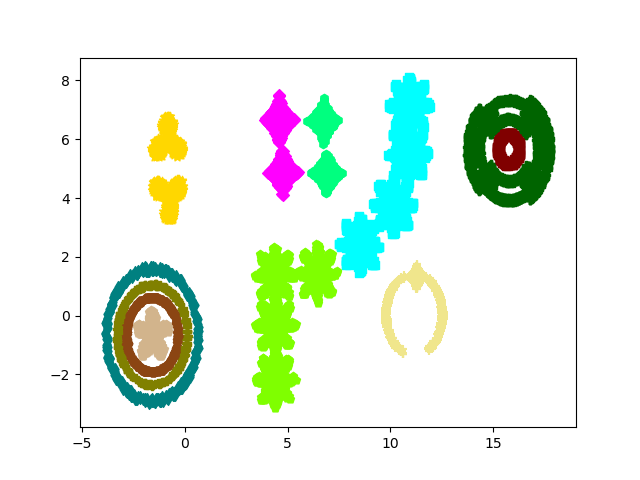}}
	
	\subcaptionbox{$\epsilon=0.4, MinPts=10$\label{fig:arti_21} }{\includegraphics[width=0.25\textwidth ]{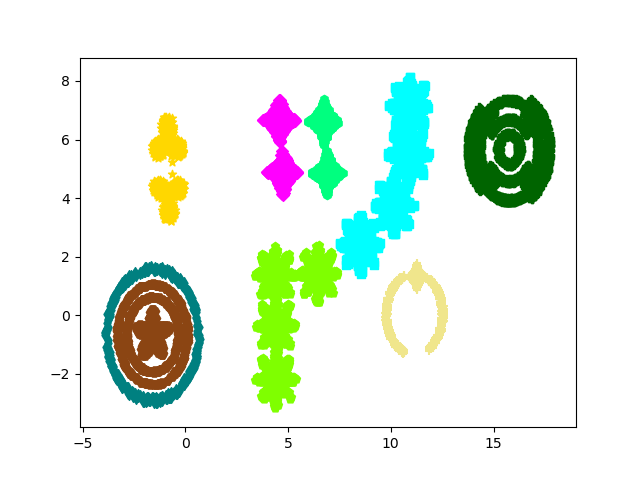}}
	
	\subcaptionbox{$\epsilon=0.6, MinPts=10$\label{fig:arti_31} }{\includegraphics[width=0.25\textwidth ]{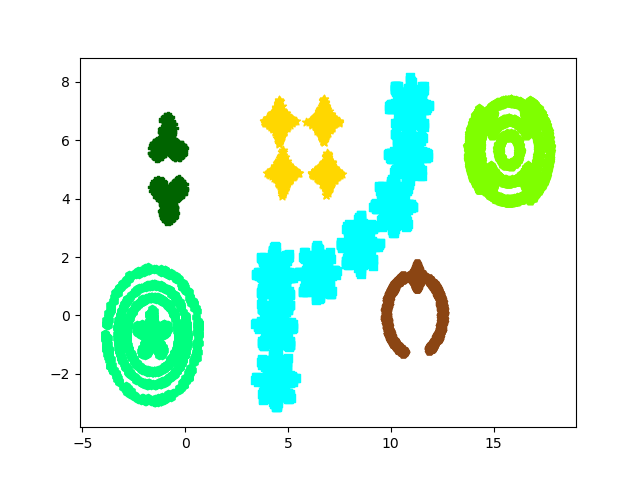}}
	
	\subcaptionbox{$\epsilon=1.0, MinPts=10$\label{fig:arti_41} }{\includegraphics[width=0.25\textwidth ]{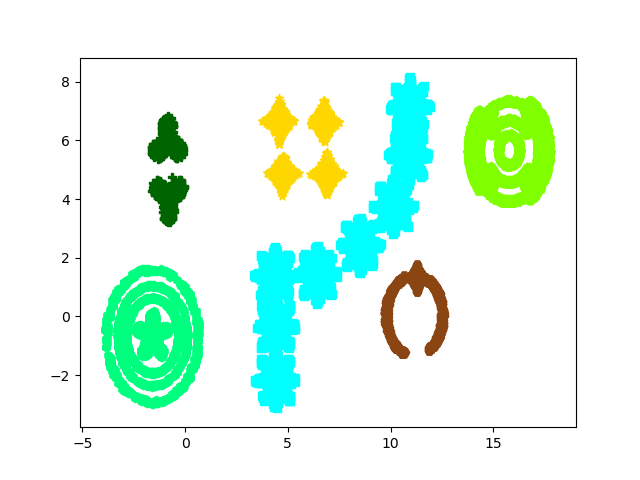}}

\end{multicols}
	
	\caption{Clustering results by (a-d) \emph{IPD} and by (e-h) \emph{IncAnyDBC} on artificial dataset.}
	\label{fig:compartificial}
\end{figure*}
\subsubsection{Analysis on {the size ($\alpha$) of test dataset}}
The size of $\alpha$ depends on both the number of clusters $k$ and the size of the dataset $n$. For a large dataset, the test size becomes significantly large.  It causes a large number of computations during the generation of test labels at every iteration. Hence, we use the size of a subset of the dataset. Here, we use $n=50,000$. Similarly, $K$ may also be a bottleneck for the system. We use the number of clusters present in the prototype for computing $\alpha$. We also fix $K$ to a reasonably high value. Here, we use $K=50$. Although this strategy eliminates the dynamic nature of the test size, it ensures scalability. 
\begin{figure*}[htb!]	
	
	\begin{multicols}{3}
		
		\subcaptionbox{iteration vs $\Delta$\label{fig:delta} }{\includegraphics[width=0.3\textwidth ]{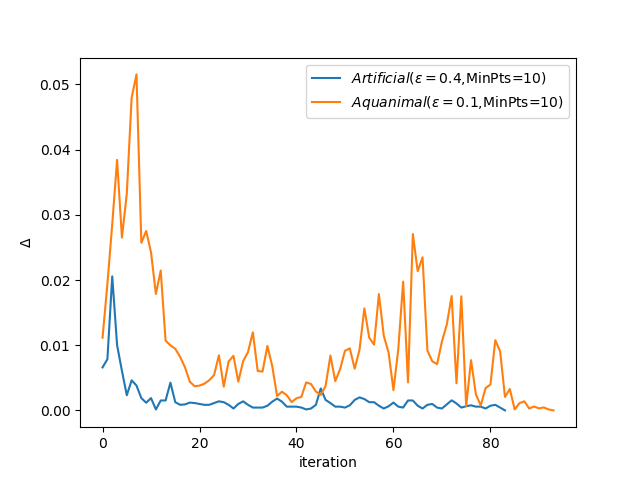}}
		
		\subcaptionbox{iteration vs K \label{fig:aqua_K} }{\includegraphics[width=0.3\textwidth ]{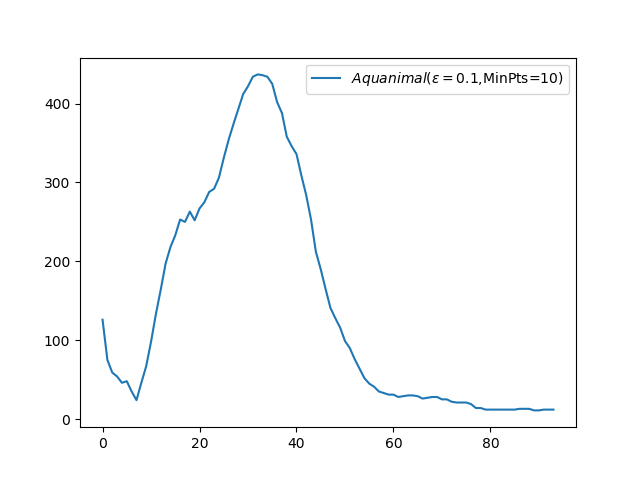}}
		
		\subcaptionbox{iteration vs K\label{fig:arti_K} }{\includegraphics[width=0.3 \textwidth  ]{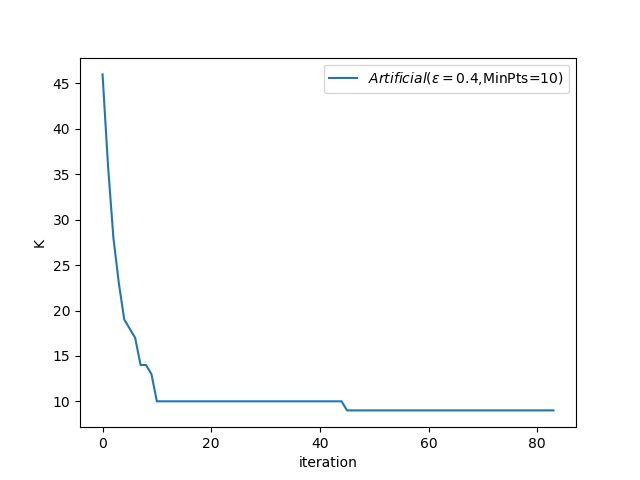}}
		
		
	\end{multicols}
	
	\caption{Convergence on large scale datasets (Aquanimal and Artificial)}
	\label{fig:convergence}
\end{figure*}
\subsubsection{Convergence Analysis}
The successful execution depends upon the fact that $\Delta$ (refer to \Cref{sec:delta}) should reach zero within fewer iteration. i.e., the method should converge after processing a fraction of the dataset such that it stops within a reasonable time. This also indicates that our method finds a stable cluster structure. Otherwise, the method stops when all data has been processed. To study the convergence, we have measured $\Delta$ and detected cluster number for each iteration.  \Cref{fig:convergence} shows respective plots for Aquanimals and Artificial datasets.  \Cref{fig:aqua_K} and \Cref{fig:arti_K} supports \Cref{lemma:converge}.


\begin{figure}[htb!]	
	
	\begin{multicols}{2}
		
		\subcaptionbox{$\tau$ vs $NMI$\label{fig:nmi} }{\includegraphics[width=0.5\textwidth ]{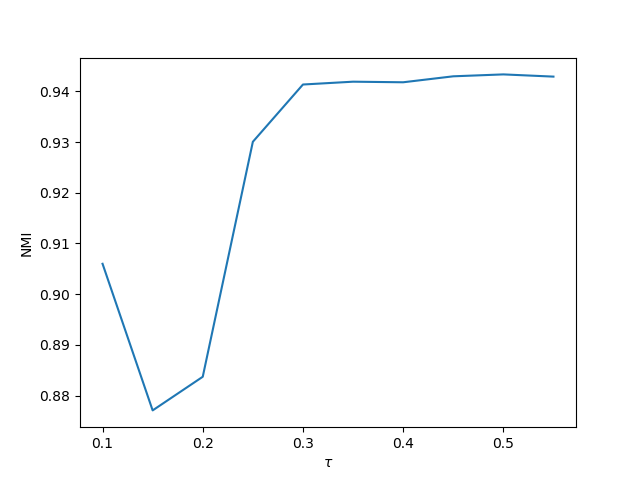}}
		
		\subcaptionbox{$\tau$ vs $runtime$ \label{fig:runtime} }{\includegraphics[width=0.5\textwidth ]{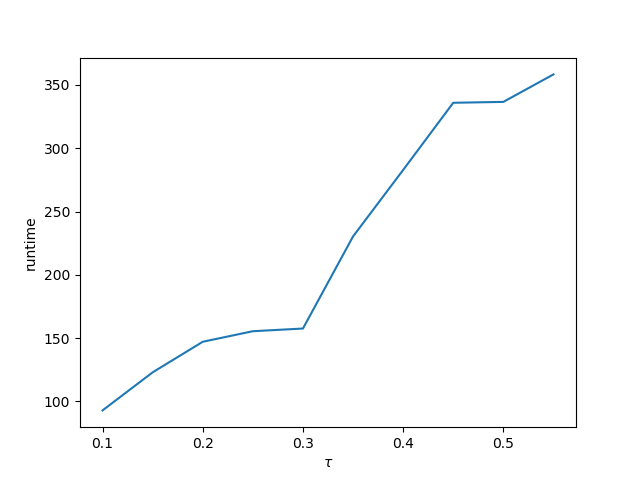}}
		
	\end{multicols}
	
	\caption{Effect of $\tau$ on IPD for Aquanimal}
	\label{fig:runtime_vs_nmi}
\end{figure}
\subsubsection{{The effect of threshold $\tau$ to select representatives}}
\begin{figure*}[htb!]	
	
	\begin{multicols}{5}
		
		\subcaptionbox{$\tau=0.1$\label{fig:tau1} }{\includegraphics[width=0.2\textwidth ]{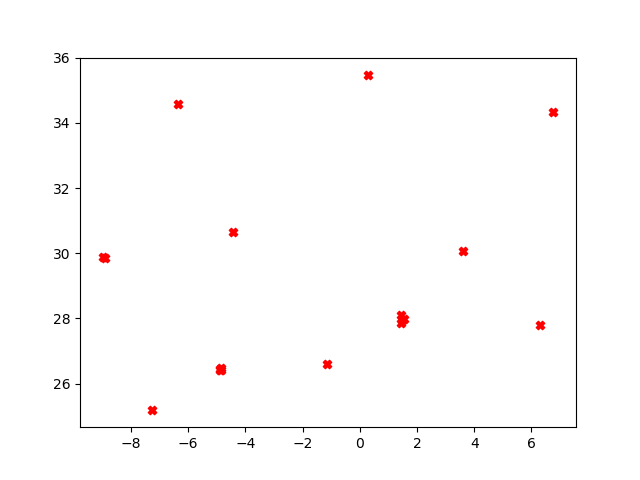}}
		
		\subcaptionbox{$\tau=0.2$\label{fig:tau2} }{\includegraphics[width=0.2\textwidth ]{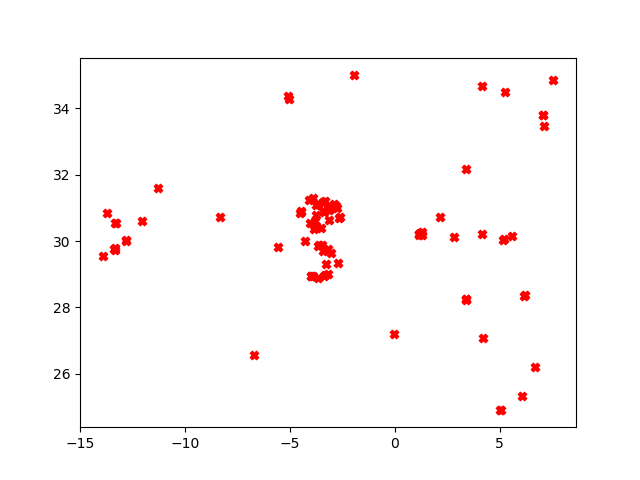}}
		
		\subcaptionbox{$\tau=0.3$\label{fig:tau3} }{\includegraphics[width=0.2 \textwidth  ]{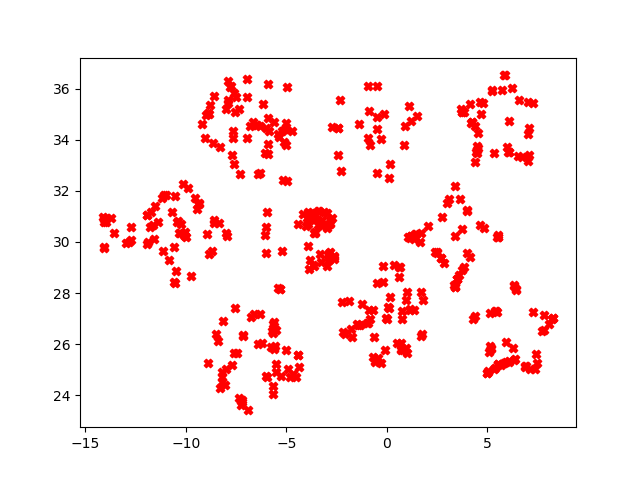}}
		
		\subcaptionbox{$\tau=0.4$\label{fig:tau4} }{\includegraphics[width=0.2 \textwidth  ]{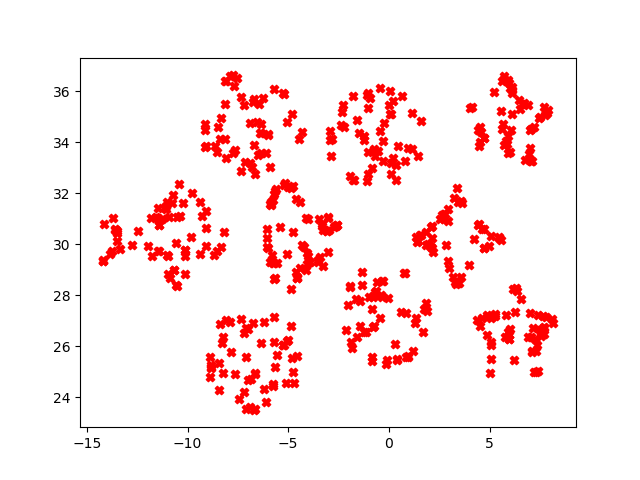}}
		
		\subcaptionbox{$\tau=0.5$\label{fig:tau5} }{\includegraphics[width=0.2 \textwidth  ]{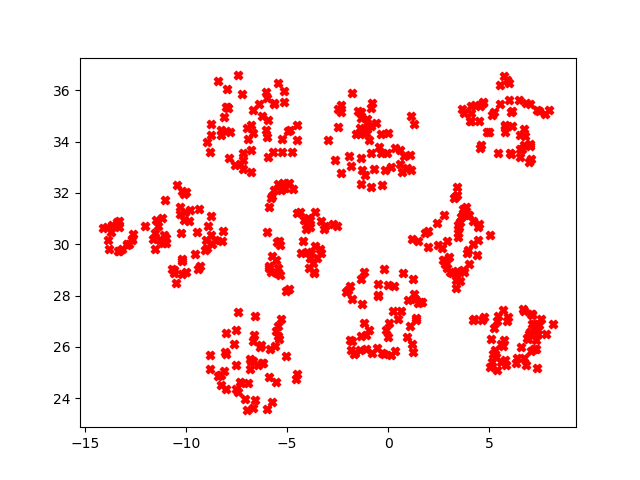}}	
	\end{multicols}
	
	\caption{Selected representative for Aquanimal at several threshold ($\tau$) }
	\label{fig:representatives}
\end{figure*}

$\tau$ is an important parameter that helps to select the number of representatives from each cluster. With the higher value of $\tau$, the number of representatives increases. But, this creates a bottleneck while handling large-scale datasets. \Cref{fig:runtime_vs_nmi} depicts that NMI reaches stability at $\tau=0.3$.  NMI does not change significantly with $\tau>0.3$. However, with increasing $\tau$, run time increases. \Cref{fig:representatives} depicts that with high increasing $\tau$, representatives can draw the contour of the clusters.  The number of representatives controls the quality of the cluster. It is a trade-off between the number of representatives and execution time.
\section{Conclusion}
{In this paper, we introduce the prototype based incremental \emph{DBSCAN} clustering algorithm, called \emph{IPD},  with the aim to select representatives for the arbitrary shaped clusters.
Our method is capable to handle large scale dataset. 
\noindent
The principle of this method is quite different from the existing DBSCAN based methods which tries to provide a scalable solution. We understand the strength of representatives of clusters in real time situation. 
\noindent
In general, IPD uses the following strategy: i) it creates prototype, ii) assign a cluster structure, iii) ask for feedback and iv)modify the prototype and its cluster structure.  The algorithm iterates over these steps until there is no feedback.\\
\noindent 
For data clustering, \emph{IPD} chooses a subset of data points to build stable cluster structure that fits to the original dataset. Hence, it consumes a fewer queries. On the other hand, obtaining a good quality of clusters  depends upon the choice of DBSCAN parameters. 
If a particular combination of $\epsilon$  and $MinPts$ identifies only one cluster in consecutive two iterations, the execution will eventually stop. We have tested with the large datasets for several combinations of $\epsilon$ and $MinPts$. We observe that {\emph IPD} is very fast Compared to {\emph IncAnyDBC}  for a few combinations of  $\epsilon$ and $MinPts$. In most cases, this situation occurs when $K$ reaches 1. The incident suggests that the convergence rate depends upon the value of $K$. However, this is not the case for \emph{IncAnyBC}. On the other hand, we observe that  \emph{IncAnyDBC} and {IPD} do not produce a similar cluster structure for the same $\epsilon$ and $MinPts$.
\noindent
Experiments suggest that \emph{IPD} can capture more than one cluster structure present in the dataset.}
	
	\section{Acknowledgments}
The authors would like to thank S. Mai for sharing their implementation of IncAnyDBC~\cite{IncAnyDBC2020}.

	\section{References}
	\bibliographystyle{model5-names}
	\bibliography{ipd.bib}

\end{document}